\documentclass[10pt,twocolumn,letterpaper]{article}

\usepackage{iccv}
\usepackage{times}
\usepackage{epsfig}
\usepackage{graphicx}
\usepackage{amsmath}
\usepackage{amssymb}

\usepackage{kotex}
\usepackage{adjustbox}
\usepackage[table, dvipsnames]{xcolor}
\usepackage{color, colortbl}
\usepackage{multirow}
\usepackage{makecell}

\usepackage{relsize}
\usepackage{textcomp}
\usepackage{multirow}
\usepackage{booktabs}
\usepackage{float}
\usepackage{adjustbox}
\usepackage{xurl}

\usepackage[pagebackref=true,breaklinks=true,letterpaper=true,colorlinks,bookmarks=false]{hyperref}

\iccvfinalcopy 

\def\iccvPaperID{8795} 

\ificcvfinal\fi

\begin{document}

\title{SCOB: Universal Text Understanding via Character-wise Supervised \\Contrastive Learning with Online Text Rendering for Bridging Domain Gap}

\author{Daehee Kim$^{1\dagger}$, Yoonsik Kim$^{1\dagger*}$, DongHyun Kim$^{1}$, Yumin Lim$^{2}$, Geewook Kim$^{1}$, Taeho Kil$^{1}$ \\
\small{$^1$ NAVER Cloud AI}
\small{$^2$ Seoul National University}\\
\tt\small{\{daehee.k,yoonsik.kim90\}@navercorp.com}
}

\maketitle
\ificcvfinal\thispagestyle{empty}\fi

\newcommand{\myparagraph}[1]{\vspace{4pt}\noindent{\bf #1}}
\renewcommand\thefootnote{\fnsymbol{footnote}}
\renewcommand{\etal}{\textit{et al.}}
\newcommand{\working}[2][blue]{\textcolor{#1}{[#2 Working]}}
\newcommand{\blue}[2][blue]{\textcolor{#1}{#2}}
\newcommand{\red}[2][red]{\textcolor{#1}{#2}}

\definecolor{darkergreen}{RGB}{21, 152, 56}
\definecolor{iblue}{rgb}{0.0,0.44,0.75}
\definecolor{pp}{rgb}{0.3,0.0,0.6}
\newcommand{\xmark}{\ding{55}}%

\begin{abstract}
Inspired by the great success of language model (LM)-based pre-training, recent studies in visual document understanding have explored LM-based pre-training methods for modeling text within document images. Among them, pre-training that reads all text from an image has shown promise, but often exhibits instability and even fails when applied to broader domains, such as those involving both visual documents and scene text images. This is a substantial limitation for real-world scenarios, where the processing of text image inputs in diverse domains is essential. In this paper, we investigate effective pre-training tasks in the broader domains and also propose a novel pre-training method called SCOB that leverages character-wise supervised contrastive learning with online text rendering to effectively pre-train document and scene text domains by bridging the domain gap. Moreover, SCOB enables weakly supervised learning, significantly reducing annotation costs. Extensive benchmarks demonstrate that SCOB generally improves vanilla pre-training methods and achieves comparable performance to state-of-the-art methods. Our findings suggest that SCOB can be served generally and effectively for read-type pre-training methods. The code will be available at \href{https://github.com/naver-ai/scob}{https://github.com/naver-ai/scob}.
\end{abstract}

\footnote[0]{$^*$Corresponding author: yoonsik.kim90@navercorp.com}
\footnote[0]{$^\dagger$Equal contribution}
\footnote[0]{$^{2}$This work was done while the author was at NAVER Cloud AI}
\section{Introduction}
\label{sec:intro}
\begin{figure}[t]
\begin{center}
   \includegraphics[width=0.85\linewidth]{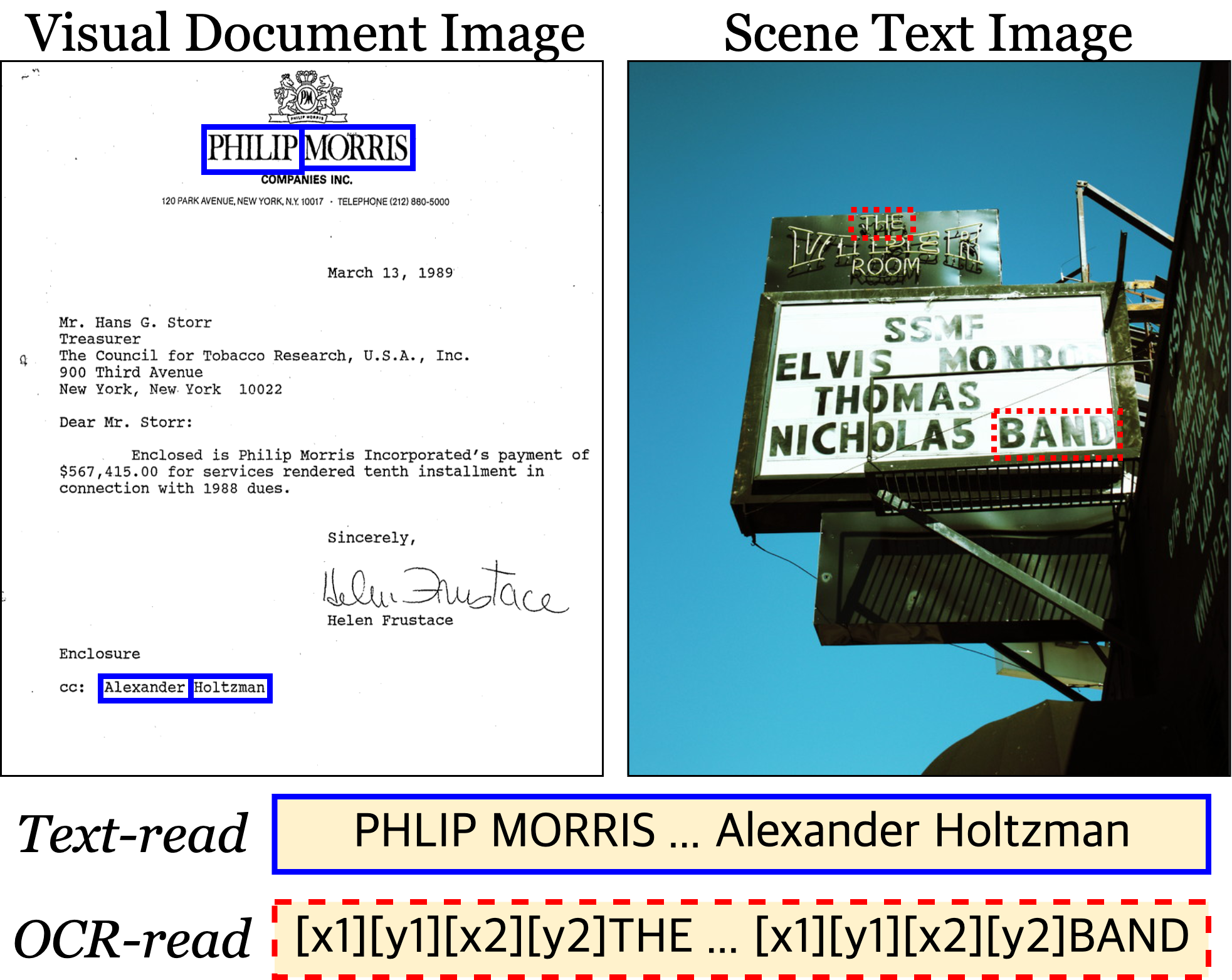}\vspace{-0.5em}
\end{center}
  \caption{\textit{Text-read} reads the text present in an image in raster scan order (blue solid box), while \textit{OCR-read} detects both the text and its corresponding coordinates from the image (red dashed box).}
\label{fig:domains}\vspace{-1em}
\end{figure}

\textit{Visually-situated language}, which encompasses a mixture of texts and visual objects such as documents, tables, infographics, and user interfaces, is now ubiquitous in modern human civilization.
Accordingly, automatically reading and understanding visually-situated language with machine learning systems is considered commercially valuable and challenging. Considering the usability and training convenience for machine learning systems, Visual Document Understanding (VDU) and Scene Text Understanding (STU) tasks have been separately studied for visually-situated language.
VDU mainly handles visually scanned or binarized document images, whereas STU takes images in real-world and dynamic environments as input, as shown in Figure~\ref{fig:domains}.

\begin{figure}[t]
\begin{center}
   \includegraphics[width=0.85\linewidth]{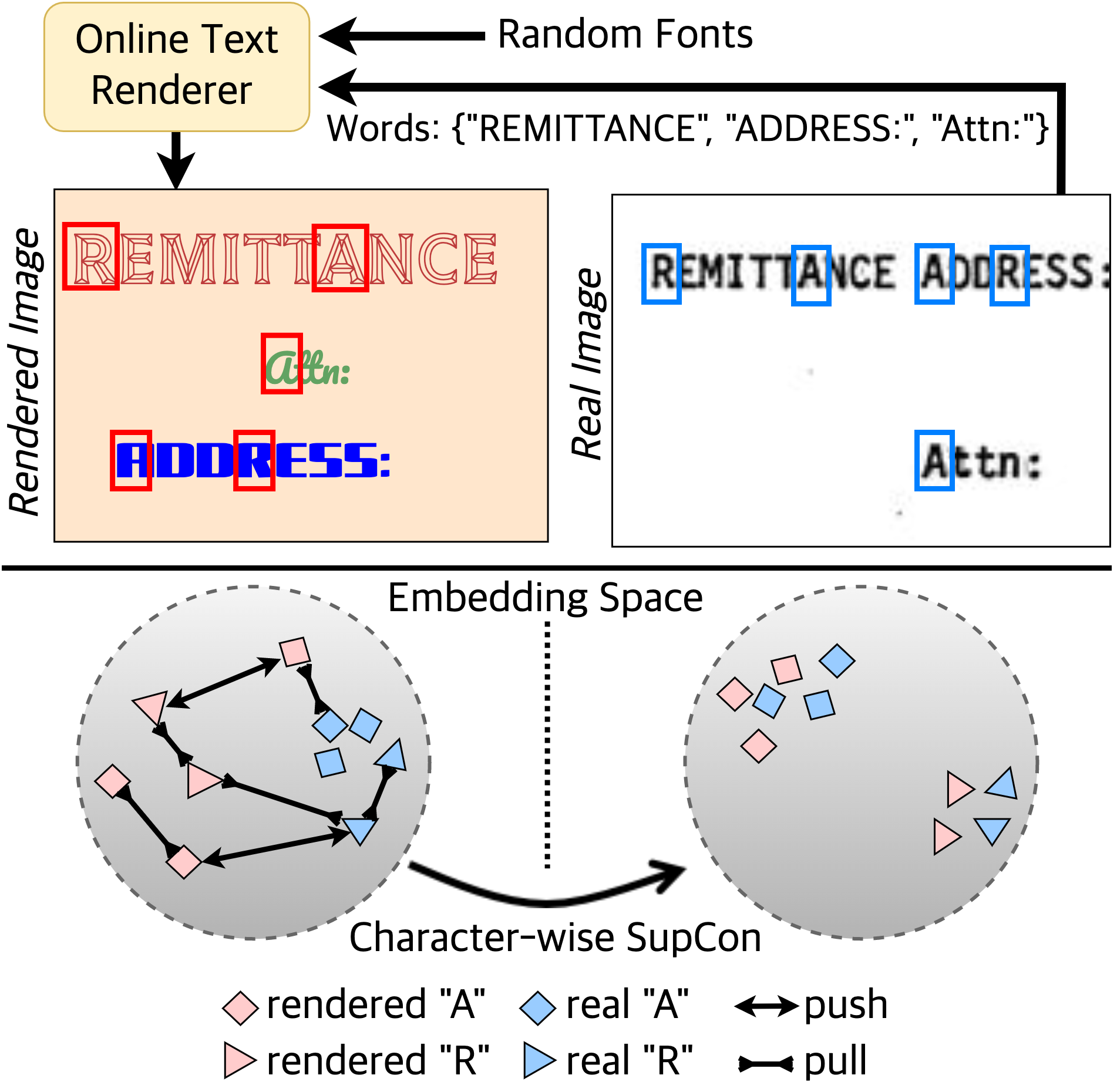}\vspace{-0.5em}
\end{center}
  \caption{Our proposed SCOB is applicable to pre-training tasks of generative text understanding models, including \textit{text-read} and \textit{OCR-read}. (Top) During the pre-training, our renderer generates images at the word-level with diverse fonts and sizes. Shapes of ``A''s and ``R''s are little different respectively in the real image (blue box), but their shapes vary significantly in the rendered image (red box). (Bottom) By applying the character-wise supervised contrastive loss, ``A''s and ``R''s are clustered respectively and the clusters of ``A'' and ``R'' push away each other in the embedding space.
  }
\label{fig:teaser}\vspace{-1em}
\end{figure}

In the context of VDU, Donut~\cite{kim2021donut} has been proposed as a sequence generation model, which pre-trains a \textit{text-read} task of reading all texts in raster scan order from an image, as illustrated in Figure~\ref{fig:domains}.
Meanwhile, Pix2seq~\cite{chen2021pix2seq} is an image-to-sequence model that extends to the object detection task by gridding images and using coordinate tokens on the grid.
These recent studies~\cite{chen2021pix2seq, kim2021donut, lu2022unified} suggest that prompt control in a sequence generation approach can  successfully expand tasks or domains more easily.
Inspired by these approaches, we investigate using VDU and STU data together for integrating text-related tasks. 

This work aims to pre-train a universal text understanding model for both document and scene domains and extend its use to downstream tasks. However, we empirically observed that \textit{text-read} using both VDU and STU data often fails and becomes unstable, likely due to the complex natural scene backgrounds in STU conflicting with the document images in VDU. To mitigate this issue, we explore \textit{OCR-read} (i.e., optical character recognition)~\cite{peng2022spts}, which explicitly guides the model to recognize text in complex images by adding coordinate tokens from \textit{text-read} in the sequence generation architecture. While \textit{OCR-read} has successfully pre-trained both domains, it requires high-cost annotation due to the need of location information, unlike \textit{text-read}. According to Yair~\etal~\cite{kittenplon2022towards}, adding box coordinate annotations for the OCR task increases annotation time by about 140\% compared to text-only annotations.

To address the issues, we propose a novel pre-training method called \textbf{SCOB}, which stands for Character-wise \textbf{S}upervised \textbf{C}ontrastive Learning (SupCon) with \textbf{O}nline Text Rendering for \textbf{B}ridging Domain Gap. 
Our online text renderer serves as a more effective augmentation method than augmentation methods used in conventional representation learning~\cite{grill2020bootstrap, he2020momentum, chen2020simple} for character-wise SupCon, as shown in Figure~\ref{fig:teaser}. 
SCOB bridges the domain gap by learning recognition more easily through rendered images and attracting the projection of positive samples from synthetic, scene text, and document domains to each other.
Applying SCOB to \textit{text-read} provides learning stability, indicating that SCOB effectively bridges the domain gap.
Moreover, pre-training \textit{OCR-read} with SCOB requires image data with only text annotation, dramatically reducing annotation costs compared to traditional OCR training. 

The proposed SCOB is applicable to a broad range of existing Transformer-based generative text understanding models, including Donut~\cite{kim2021donut}, Dessurt~\cite{davis2022end}, and Pix2Struct~\cite{lee2022pix2struct}. From a generalized perspective, these image-to-text models can be interpreted as the same framework, which is a transformer-based encoder-decoder model with a ``read'' pre-training strategy (e.g., \textit{text-read} and \textit{OCR-read}). We refer to this framework as the \textbf{U}niversal Text \textbf{U}nderstanding (\textbf{W}) framework in this work. We conduct extensive experiments using the \textbf{W} framework on eleven benchmarks spanning both VDU and STU domains to observe the characteristics and effects of respective pre-training strategies with SCOB. 
Our experimental results and analysis demonstrate the efficacy and versatility of SCOB improving the overall model performance.
We summarize our main contributions as follows:\vspace{-0.5em}
\begin{itemize}
    \item This paper investigates the effects of \textit{text-read} and \textit{OCR-read} pre-training on a total of eleven tasks, including those in the VDU and STU domains. \vspace{-0.5em}   
    \item We propose a novel pre-training method SCOB that utilizes character-wise contrastive learning with online text rendering to effectively bridges the domain gap between VDU and STU domains.  \vspace{-0.5em}
    \item SCOB enables weakly supervised OCR pre-training, significantly reducing annotation costs by using only text annotations. \vspace{-0.5em}
    \item 
    Experimental results demonstrate the effect of read-based pre-training for a table reconstruction task, achieving state-of-the-art. Moreover, our proposed SCOB generally enhances the performance of read-based pre-training on various text-related downstream tasks.
    \vspace{-0.5em}
\end{itemize}

\section{Related Work}
\label{sec:related}

\begin{figure*}[t]
\begin{center}
   \includegraphics[width=0.9\textwidth]{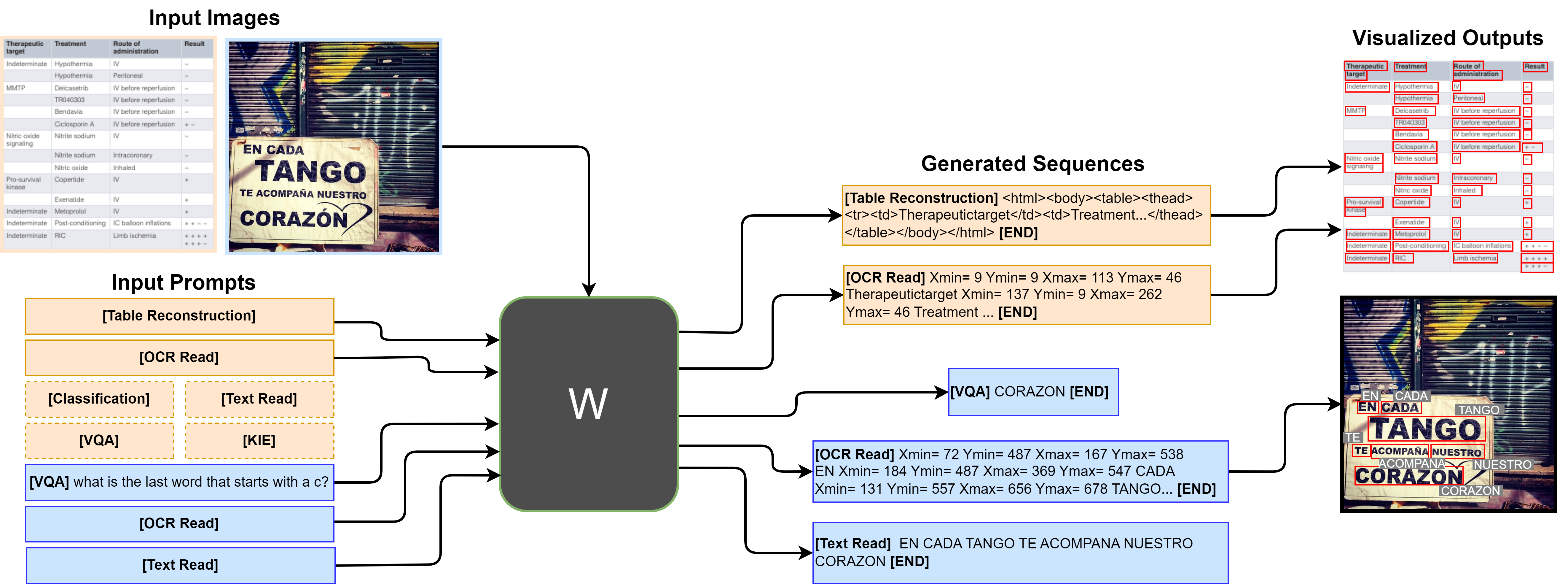}
\end{center}
   \caption{
   The overview of the \textbf{U}niversal Text \textbf{U}nderstanding (\textbf{W}) framework. This framework provides a unified approach for various visual text-related tasks. Given an input text image, \textbf{W} generates output sequences for downstream tasks conditioned on task-prompts in the text decoder. We aim to train the \textbf{W} framework with pre-training tasks such as \textit{text-read} or \textit{OCR-read} to effectively handle both VDU (yellow box) and STU (blue box) domains. When the model is pre-trained without domain conflicts, it can be fine-tuned on various text understanding tasks spanning both domains, such as table reconstruction, OCR, classification, VQA, and KIE, with improved performance.}
   \label{fig:multitask}\vspace{-1.0em}
\end{figure*}

\subsection{Visual Document Understanding}
Inspired by the great success of BERT~\cite{kenton2019bert} in natural language process tasks, Xu \etal \cite{xu2020layoutlm} presented a powerful VDU model, LayoutLM, with an efficient pre-training task, named masked visual-language modeling.
Recently, LayoutLMv3~\cite{huang2022layoutlmv3} exploited masked image modeling with latent codes of a discrete VAE and achieved state-of-the-art.
However, these approaches~\cite{xu2020layoutlm, xu2021layoutlmv2, li2021selfdoc, huang2022layoutlmv3, hong2022bros} require a specific architectural design for the output format of each downstream task.
Moreover, since they use OCR results as input, they strongly depend on the OCR engine.
In addition, OCR increases overall computational cost, and the errors of OCR often propagate to the final outputs~\cite{kim2021donut}.

In order to solve these challenges, Kim~\etal~\cite{kim2021donut} proposed Donut that does not require preprocessing as OCR.
Donut is an end-to-end encoder-decoder model that auto-regressively generates the desired type of output sequence.
With a simple concept, Donut solves multiple VDU tasks with a single unified pipeline, and it showed state-of-the-art performances on various VDU tasks.
Donut is pre-trained with a task, denoted as \textit{text-read} task, which is simply reading all characters in the image with raster order~\cite{kim2021donut}.
Although Donut showed promising results on many VDU tasks, it has not been investigated yet how it performs in scene text-related tasks.
In addition, we note that text localization is likely to be important for some text-related tasks, which have not been explored deeply in the previous works~\cite{kim2021donut,davis2022end}.
In this paper, \textit{text-read} and \textit{OCR-read} are investigated as pre-training methods and their impact on VDU and STU downstream tasks.

\subsection{Contrastive Learning for Visual Representation Learning}
In the computer vision, unsupervised representation learning methods have succeeded with contrastive learning~\cite{grill2020bootstrap, he2020momentum, chen2020simple, chen2020improved, khosla2020supervised, chen2021empirical}. 
The common idea of these methods is that image augmentation is performed on a single batch of images, where a pair of augmented images can be treated as a positive pair originally taken from the same image and a negative pair originally taken from two different images.
The augmentation plays a critical role in contrastive learning, and the effect of its type and intensity has been extensively investigated~\cite{grill2020bootstrap, he2020momentum, chen2020simple, chen2020improved,  jing2022understanding}.
Another critical factor is a large number of samples~\cite{chen2020simple}; thus, dictionary-based methods~\cite{wu2018unsupervised, he2020momentum, chen2021empirical} have been proposed to cache the negative samples. 
To fully leverage supervision, supervised contrastive learning methods have been proposed~\cite{khosla2020supervised, islam2021broad} where the positives and negatives are constructed with their label information.

STU field, especially OCR, has also explored the application of contrastive learning to train image and text encoders~\cite{Baek_2021_CVPR, aberdam2021sequence, xue2022language}. 
Specifically, self-supervised learning was employed for text recognizer~\cite{ Baek_2021_CVPR, aberdam2021sequence} and these methods can be mainly categorized by the instance-level contrastive learning.
Baek~\etal~\cite{Baek_2021_CVPR} defined instance-level as an image and applied MoCo~\cite{he2020momentum} for representation learning. 
On the other hand, Aberdam~\etal~\cite{aberdam2021sequence} defined instance-level as a sub-image under an assumption that the placement of texts in positives would not be that different unless severe placement-related augmentations (e.g., flip) are applied.
Recently, CLIP-based~\cite{radford2021learning} contrastive learning was proposed to train both text and image encoders with label information. 
In this paper, we introduce contrastive learning for representation learning, which differs from previous methods. 
We pre-train the auto-regressive text decoder as well as the image encoder with SCOB, which can be effectively transferred to downstream tasks.

\section{Method}
\label{sec:method}

\begin{figure*}[t]
\begin{center}
   \includegraphics[width=0.78\textwidth]{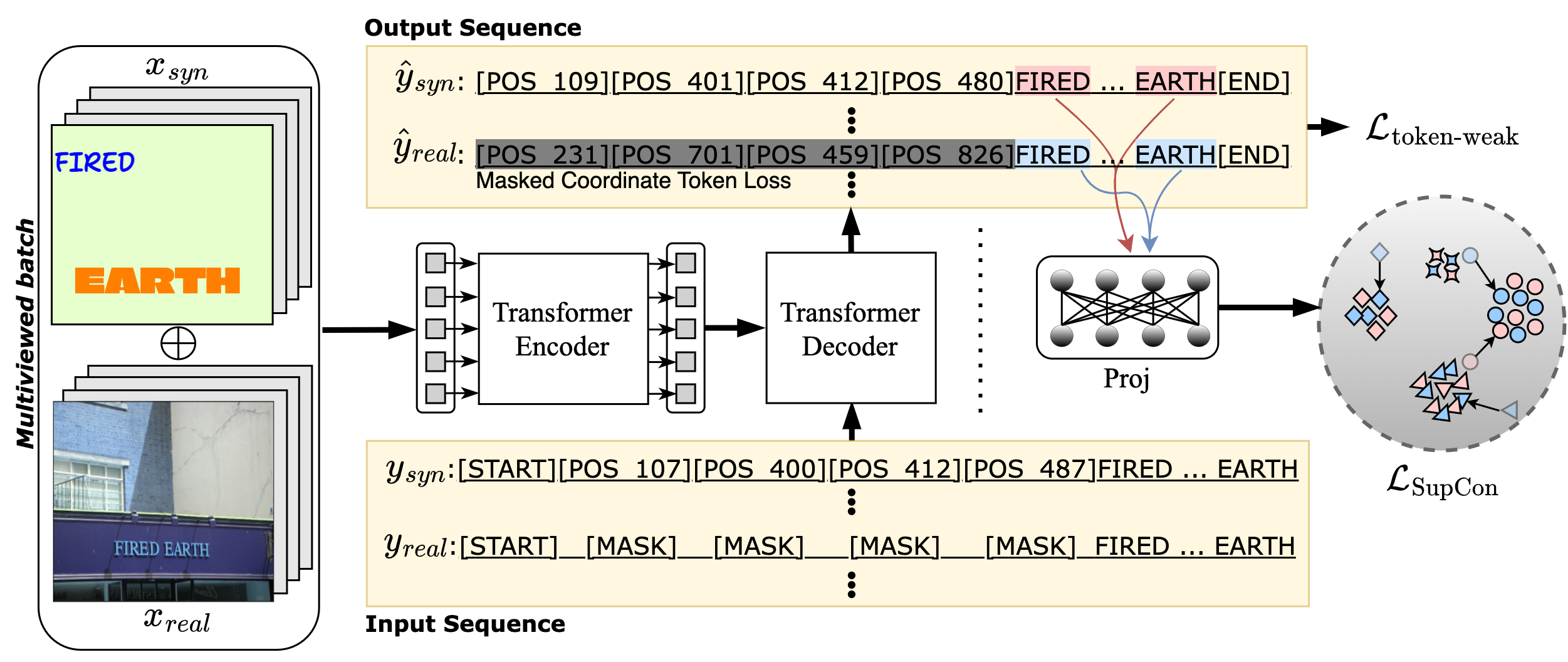}\vspace{-0.5em}
\end{center}
   \caption{
    An illustration of the proposed \textbf{SCOB}-applied \textit{OCR-read} pre-training method for text understanding models. 
    SCOB can also be applied to \textit{text-read} by excluding the coordinate tokens.
    Given real images, $\mathbf{x}_{real}$, our renderer generates a corresponding synthetic batch, $\mathbf{x}_{syn}$, and the multiviewed batch is used to train the model. 
    Under a teacher-forcing scheme~\cite{williams1989learning}, the model is trained using a cross-entropy loss, $\mathcal{L}_\textnormal{token-weak}$, along with a contrastive loss, ${\mathcal{L}}_\textnormal{SupCon}$. 
    Notably, SCOB does not require coordinate annotations of $\mathbf{y}_{real}$. The coordinate labels of $\mathbf{y}_{real}$ are replaced with masked tokens ([MASK]), and the loss from the masked tokens is ignored (gray box).
    To compute the ${\mathcal{L}}_\textnormal{SupCon}$, the predicted character embeddings are passed through an MLP projector ($\text{Proj}$). 
    Note that the predicted character embedding is the last layer hidden embedding of the decoder, $\mathbf{d}$, not the character token.
    With SCOB, the same classes of characters are forced to be clustered in the embedding space, leading to improved model robustness.
    } 
   \vspace{-1.em}
   \label{fig:main_supcon}
\end{figure*}

Inspired by Donut~\cite{kim2021donut}, we adopt the sequence generation model to process various downstream tasks with a single architecture. For pre-training the sequence generation model, we investigate two objectives: \textit{text-read} and \textit{OCR-read}. As shown in Figure~\ref{fig:multitask}, \textit{text-read} is simply reading all characters in the image with raster order~\cite{kim2021donut}. \textit{OCR-read} incorporates \textit{text-read} and text localization objectives that decodes the coordinates of bounding boxes and text transcriptions. 
We expect \textit{OCR-read} can employ richer information packaging physical coordinates and sizes, as well as the relative distances between text instances~\cite{yang2022unitab}. Since text localization occupies part of the target sequence, shorter text transcriptions can be exploited for a learning language model because the decoder has a limited decoding max length. Thus, \textit{text-read} learns a language model more comprehensively when the image contains many text instances that the decoding length of \textit{OCR-read} cannot cover entire text instances.

Furthermore, we propose SCOB, a novel pre-training approach that leverages character-wise SupCon and online text rendering to maximize their synergy. The online text renderer serves as a suitable augmentation method for general text-related contrastive learning. Document images~\cite{lewis2006building} are usually well-scanned or binarized, making it easy for models to recognize text, whereas scene text images can be challenging due to the natural background. 
Also, we observe that \textit{text-read} is not robust enough to pre-train document and scene text data together (see Section~\ref{subsec:eval}).
SCOB overcomes this limitation by training the model on synthetic text images, which are easier to recognize, and transferring this knowledge to more challenging real-world scenarios. Specifically, SCOB pulls the feature of positive samples from synthetic, document, and scene text images, facilitating the learning of difficult samples. Moreover, SCOB supports \textit{text-read} and \textit{OCR-read} pre-training methods and can be trained with weak supervision for \textit{OCR-read}, making it an ideal solution for scenarios where data is scarce.

In the next Section~\ref{ss:baseline}, we explain the architecture of \textbf{W} and vanilla pre-training objectives. Then, we discuss the character-wise supervised contrastive learning method pre-training in Section~\ref{ss:supcon}. 
Finally, Section~\ref{ss:otor} describes the detailed settings of the online text renderer and the weakly supervised pre-training method. 

\subsection{Read-based Pre-training}
\label{ss:baseline}
Recently, the proposal of Pix2Seq~\cite{chen2021pix2seq} made detection possible with sequence generation, which allows unifying the output of multiple tasks, including detection. Our \textit{OCR-read} pre-training is also in the form of sequence generation, and the sequence is composed of coordinates (bounding box) and transcription as shown in Figures~\ref{fig:multitask}, and~\ref{fig:main_supcon}. To express the bounding box as a sequence, we uniformly discretize the height and width of the image into 1,000 bins following Pix2Seq. Therefore, the sequence of word instance consists of 4 coordinate tokens [$x_{min}$, $y_{min}$, $x_{max}$, $y_{max}$], followed by $n$ character tokens (transcription). 
In the case of \textit{text-read} pre-training, the target sequence is composed of only character tokens, which can handle more words than \textit{OCR-read}.

\myparagraph{Architecture.}
The architecture of $\textbf{W}$ follows the encoder-decoder framework.
The image encoder converts the input image $\mathbf{x}{\in}\mathbb{R}^{H\times W\times C_{in}}$ into visual embedding $\mathbf{v}$ = Enc($\mathbf{x}$) where $H$, $W$, and $C_{in}$ denote the height, width, and channel of the input image, respectively.
The text decoder takes both $\mathbf{v}$ and previously generated token from the decoder as input and auto-regressively generates output sequence $(\mathbf{\hat{y}})_{i=1}^N$ where  $\mathbf{\hat{y}}_i$ is the $i$-th generated token, and $N$ denotes the sequence length of the decoder. 
We use Swin Transformer~\cite{Liu_2021_ICCV} and Transformer-based~\cite{vaswani2017transformer} decoder as an encoder and decoder, respectively.

\myparagraph{Objective.}
The model learns to predict target tokens such as prompt, coordinate (only for \textit{OCR-read}), and character tokens, using maximum likelihood:
\begin{align}\label{math:objective}
    \mathcal{L}_\textnormal{token} = -\sum_{i=1}^N\log P(\mathbf{y}_i|\mathbf{x},\mathbf{\hat{y}}_{1:i\text{-}1}), 
\end{align}
where $\mathbf{y}_i$ denotes $i$-th target token. 

\subsection{Character-wise Supervised Contrastive Loss}
\label{ss:supcon}
In general image classification, supervised contrastive learning~\cite{khosla2020supervised} has been mostly applied at image-level.
On the other hand, in the image containing text, each character can be regarded as an instance and we propose character-wise contrastive learning. 
The proposed method enables stable learning without using a kind of memory bank (dynamic dictionary queue~\cite{he2020momentum}) because character-wise SupCon allows obtaining abundant positive and negative samples even in a small batch where most OCR images have high-resolution sizes (larger than 768$\times$768).
Another major factor of contrastive learning is the augmentation for multiview images. 
We propose to generate multiview images using the online text renderer, which will be described in the following subsection.

Figure~\ref{fig:main_supcon} shows the overview of the proposed character-wise SupCon. 
Our model takes original (real) and multiview (synthetic) images as the input and auto-regressively generates the token $\hat{y}_i=\text{MLP}({\mathbf{d}_i})$ where $\mathbf{d}_i$ denotes the last hidden embedding of the decoder at $i$-th generation index. 
At the same time, character-wise projections $\mathbf{z}_i = \text{Proj}({\mathbf{d}_i})$ are placed in a contrastive subspace. 
Here, we define a multiviewed batch as a union of original batch and rendered batch constructed by the renderer.
Within a multiviewed batch, let $j \in \mathbf{C}$ be the index of a character where $\mathbf{C}$ denotes the set of all target characters in multiviewed batch.  
Then, the model is trained with character-wise supervised contrastive loss: 
\begin{align} \label{eq:supcon}
    \mathcal{L}_\textnormal{SupCon} =  \sum_{j\in \mathbf{C}}\frac{-1}{|P(j)|} \sum_{p \in P(j)}\log\frac{\exp(\mathbf{z}_j\cdot\mathbf{z}_p/\tau)}{\sum_{a \in A(j)}\exp(\mathbf{z}_j\cdot\mathbf{z}_a/\tau)}, 
\end{align}
where $A(j) = \mathbf{C}\setminus\{j\}$, and $P(j) = \{p \in A(j): \mathbf{c}_p=\mathbf{c}_j\}$ is the set of indices that have same character label $\mathbf{c}$ in the multiviewed batch.
$|P(j)|$, symbol $\cdot$ and $\tau$ denote cardinality of $P(j)$, dot product, and scalar temperature, respectively.

\subsection{Online Text Renderer}
\label{ss:otor}
To address the limitations of existing augmentation methods for OCR in contrastive learning~\cite{Baek_2021_CVPR,aberdam2021sequence,radford2021learning}, we propose the online text renderer. Existing methods cannot employ strong geometric augmentations like crop and flip, as the same word must be identically contained in multiview images, which weakens the variance between the positive views. In contrast, the renderer generates an image corresponding one-to-one with the original image using text transcription, outputting the synthetic image and the bounding box annotation together. This creates a more substantial variance between the same characters even without strong augmentation, such as cropping.

The renderer is designed specifically for the character-wise SupCon and differs from existing renderers~\cite{Gupta16,kim2021donut} in two significant ways:
i) We adopt an online generation method to replace existing augmentation in contrastive learning.
ii) To maximize the character variance with high speed, we randomly select various fonts and background colors.
We also observed that the renderer could generate synthetic data in less time than it takes to load real data into memory. Overall, the proposed renderer presents a simple and efficient approach to text rendering that can improve the variance between positive views, leading to better contrastive learning results.

\myparagraph{Online Rendering Engine.}
Our generation engine is implemented using the Python Pillow package~\cite{clark2015pillow}. To generate synthetic text images, we require only a font and a set of words. We leverage more than 3,000 fonts provided by Google\footnote{\url{https://fonts.google.com}} to maximize the character variance. Detailed settings can be adjusted, including image resolution range, background RGB range, font size range, and whether to generate character-level coordinates. The background color is chosen randomly within the specified range, and each word in the set is rendered at a randomly selected location and size within the specified range. Examples of synthetic images are shown in Figures~\ref{fig:teaser} and~\ref{fig:main_supcon}.

\myparagraph{Weakly Supervised Pre-training OCR.} 
We propose a weakly supervised pre-training for \textit{OCR-read} that eliminates the need for expensive coordinate annotation of real images. Specifically, the model learns coordinate information solely from rendered data. As shown in Figure~\ref{fig:main_supcon}, the proposed weakly supervised learning involves two steps:
i) replacing the input coordinate tokens of real data with mask tokens in a teacher-forcing scheme~\cite{williams1989learning}, and ii) masking the coordinate token loss of the real data.
As a result, the model learns localization and recognition on synthetic data while only learning text annotations on real data.
Additionally, the renderer generates the same characters of input text annotations in multiview images, providing high-quality positive samples for character-wise SupCon. The proposed weakly supervised learning can be expressed as follows:
\begin{align}\label{math:OTOR}
    \mathcal{L}_\textnormal{token-weak} = -\sum_{i=1}^{N}w_{i}\log P(\mathbf{y}_i|\mathbf{x},\mathbf{\hat{y}}_{1:i\text{-}1}), 
\end{align}
where $w_{i}$ denotes a pre-assigned weight for coordinate tokens in the sequence. 
We set $w_{i}=0$ for the coordinate tokens of real data and set $w_{i}=1$ for other cases. 
In the case of \textit{text-read}, $\mathcal{L}_\textnormal{token-weak}$ is equivalent to $\mathcal{L}_\textnormal{token}$ due to the absence of the coordinate tokens.

\myparagraph{Loss Function.}
Finally, we present the following loss function $\mathcal{L}$ of SCOB that consists of token loss $\mathcal{L}_\textnormal{token-weak}$ as well as our character-wise supervised contrastive loss $\mathcal{L}_\textnormal{SupCon}$:
\begin{equation}\label{eq:total_loss}
    \mathcal{L} = \frac{1}{M}\sum_{m=1}^M\mathcal{L}_\textnormal{token-weak}^m + \lambda{\mathcal{L}_\textnormal{SupCon}},
\end{equation}
where $M$ is the number of image-label pairs in multiviewed batch and $\lambda$ denotes a scaling factor of $\mathcal{L}_\textnormal{SupCon}$.

\section{Experiments}
\label{sec:experiments}
{
\setlength{\tabcolsep}{4pt}
\renewcommand{\arraystretch}{1.3} 
\begin{table*}[t]
	\begin{center}
    	\resizebox{1.\linewidth}{!}{%
    	\begin{tabular}{@{}lccccccc|ccccc@{}} \toprule
    	    \multirow{3}{*}{Method} & \multirow{3}{*}{\#GPUs} & \multicolumn{11}{c}{Universal Text Understanding Downstream Tasks}  \\ \cmidrule{3-13}
            
            & &   \makecell[c]{Table\\Reconstruction} & KIE & \makecell[c]{Document\\Classification} & \multicolumn{2}{c}{Document VQA}& \makecell[c]{Layout\\Analysis} & \multicolumn{3}{c}{Scene Text OCR} & \multicolumn{2}{c}{Scene Text VQA}  \\\cmidrule{3-13}
            & &    PubTabNet~\cite{zhong2020image}  & CORD~\cite{park2019cord} & RVL-CDIP~\cite{harley2015evaluation}  & DocVQA~\cite{mathew2021docvqa} & InfoVQA~\cite{mathew2022infographicvqa} & PubLayNet~\cite{zhong2019publaynet} & IC13~\cite{karatzas2013icdar} & IC15~\cite{karatzas2015icdar} & TotalText~\cite{ch2017total} & TextVQA~\cite{singh2019towards} & ST-VQA~\cite{biten2019scene}  \\ \midrule 
            W$_\textnormal{OCR-read}$ & 8$\times$V100 &  96.0  & 88.2 & 94.2 & 56.1 & 22.7 & 93.8 & 95.8 & 89.6 & 84.9 & 55.4 & \textbf{62.9}  \\
            W$_\textnormal{OCR-read}$ w/ SCOB & 8$\times$V100 & 95.9 (\blue{-0.1})  & 88.5 (\red{+0.3}) & 94.6 (\red{+0.4}) & 60.2 (\red{+4.1}) & \textbf{28.5} (\red{+5.8}) & 93.9 (\red{+0.1}) & \textbf{96.6} (\red{+0.8}) & \textbf{90.9} (\red{+1.3}) & \textbf{86.0} (\red{+1.1}) & \textbf{56.2} (\red{+0.8}) & 62.6 (\blue{-0.3}) 
            \\ \midrule
            W$_\textnormal{text-read}$ & 8$\times$V100 & \textbf{96.2} & 85.5 & 94.4 & 57.0 & 25.2 & 93.6 & 96.0 & 87.2 & 83.7 & 49.3 & 57.2 \\
            W$_\textnormal{text-read}$ w/ SCOB & 8$\times$V100 &96.0 (\blue{-0.2})	& 87.4 (\red{+1.9})	&94.3 (\blue{-0.1})	&59.6 (\red{+2.6})	&27.5 (\red{+2.3})	&93.9 (\red{+0.3})	& 96.0 (+0.0)	& 90.2 (\red{+3.0})	&85.3 (\red{+1.6})	&54.4 (\red{+5.1})	&61.2 (\red{+4.0}) 
            \\ \midrule
            TableFormer~\cite{nassar2022tableformer} & n/a & 93.7  & - & - & -& - & - & - & -  & - & - & - \\
            Donut$_\textnormal{proto}$~\cite{kim2021donut} & 8$\times$V100& - & 85.4 & 94.5 & 47.1 & 10.2 & - & - & - & - & - & -  \\
            Donut~\cite{kim2021donut} & 64$\times$A100& - & \textbf{90.9} & 95.3 & 67.5 & 24.4 & - & - & - & - & 36.8* & 61.5*  \\
            LayoutLMv3~\cite{huang2022layoutlmv3} & 32$\times$V100 &  - & 84.4* & \textbf{95.5} & \textbf{83.4} & - & \textbf{95.1} & - & - & - & - & -  \\
            SPTS~\cite{peng2022spts} & 32$\times$V100 &  - & - & - & -& - & - & 93.3 & 77.5 & 82.4 & - & -\\
            PreSTU~\cite{kil2022prestu} & n/a & - & - & - & - & -  &-&-&-&- & 54.5 & 62.6\\             
            \midrule
            UDOP~\cite{tang2023unifying}& n/a & - & - & \textbf{96.0} & \textbf{84.7} & \textbf{47.4} & - & - & - & - & - \\
            PaLI-3B~\cite{chen2023pali} & n/a & - &  - & - & - & - & - & - & - & - & 60.1 & 67.5 \\
            PaLI-17B~\cite{chen2023pali} & 1024$\times$TPUv4 & - &  - & - & - & - & - & - & - & - & \textbf{71.8} & \textbf{77.1} \\
            GIT2~\cite{wang2022git} & n/a & - &  - & - & - & - & - & - & - & - & 68.4 & 75.1 \\
            \bottomrule
        \end{tabular}}
	    \caption{The extensive benchmarks for text-related downstream tasks. ``\#GPUs'' denotes the total number of employed GPUs for pre-training. 
     The left section is VDU tasks, and the right section is STU tasks. 
     The best performance is represented in \textbf{bold}. Note that Donut was pre-trained on IIT-CDIP and SynthDoG, while  Donut$_\textnormal{proto}$ was pre-trained on SynthDoG~\cite{kim2021donut}. $*$ denotes the performance results of our fine-tuning, conducted following the author's guidelines. Note that UDOP and PaLIs employ the result of OCR as an input. The total number of training parameter of UDOP, PaLI-3B, PaLI-17B and GIT2 are 794M, 3B, 17B and 5.1B, respectively.
     }
	    \label{tab:coverage}
     \end{center}\vspace{-2.0em}
\end{table*}
}

\subsection{Pre-training Details}\label{ss:pretrain_details} \vspace{-0.5em}
\myparagraph{Architecture Setup.}
Architecture of \textbf{W} has a few changes from Donut~\cite{kim2021donut}. We use Swin-B~\cite{liu2021swin} pre-trained with ImageNet-22K~\cite{imagenet2009} as a visual encoder and set the layer numbers to [2, 2, 18, 2], window size to 12, and input resolution to 768$\times$768. For the decoder, we adopt Transformer~\cite{vaswani2017transformer} decoder initialized by BERT~\cite{kenton2019bert} and use 12-layer, 12-head, 768 hidden size, and absolute position encoding with 512 maximum sequence length. We employ a character tokenizer, as we empirically observed that character tokenizer outperforms subword tokenizer in OCR. The model is trained for 1M steps using Adam~\cite{kingma2014adam} optimizer with a batch size of 32 distributed across 8 NVIDIA V100 GPUs. 
For the training of SCOB, we set 0.5 as $\lambda$, 0.07 as $\tau$, and the character-level projector composed of the 2-layer MLP with 128 hidden dimensions.
Since SCOB for \textit{OCR-read} is performed under the assumption that the coordinates of the real dataset are unavailable, it learns a sequence constructed in random order.
The coordinate token is configured in the form of a bounding box for all datasets.

\myparagraph{Dataset.}
As a train set for pre-training, we utilize IIT-CDIP~\cite{lewis2006building} and real scene text datasets~\cite{karatzas2013icdar, karatzas2015icdar, ch2017total, OpenImages2, singh2021textocr, long2022towards}, which are widely adopted in VDU and scene text OCR, respectively. 
The batch ratio is set to 20\% and 80\% for IIT-CDIP and scene text datasets, respectively. 
IIT-CDIP is a dataset composed of 11M scanned English document images with abundant sentence-level texts and we achieve pseudo-OCR labels through the CLOVA OCR API following Donut~\cite{kim2021donut}. 
This dataset is especially useful for learning language modeling.
We employ real scene text dataset such as ICDAR2013~\cite{karatzas2013icdar}, ICDAR2015~\cite{karatzas2015icdar}, TotalText~\cite{ch2017total}, OpenImages v6~\cite{OpenImages2}, TextOCR~\cite{singh2021textocr} and HierText~\cite{long2022towards} where the total amount of data is 857K. 
For OpenImages v6, we filter non-text images and obtain pseudo-OCR labels through CLOVA OCR API.
These scene text datasets contain word instances and complex backgrounds, enabling the model to learn coordinate information and embed diverse visual features in contrastive subspace. 

\subsection{Fine-tuning Details on Downstream Tasks}
To present a comprehensive investigation, we provide extensive benchmarks on 11 datasets as shown in Table~\ref{tab:coverage}. 
Although \textit{OCR-read} has the advantage of the text localization objective, adding coordinate tokens can cause the maximum sequence length to be relatively insufficient.
To compensate for this, we perform a \textit{text-read} task of 50K as short intermediate training just before fine-tuning. 
Fine-tuned downstream tasks are briefly described as follows.
We will provide fine-tuning details in the supplemental file.

\myparagraph{Scene Text OCR.}
To evaluate the text localization objective, we fine-tune and evaluate the widely used scene text OCR datasets: ICDAR2013, ICDAR2015, and TotalText. 
The train set is the pre-training dataset excluding IIT-CDIP.

\myparagraph{Table Reconstruction.}
PubTabNet is a dataset annotated with HTML format that contains 500K training, 9K validation, and 9K test samples. 
In this paper, our models decode contents in the cell as well as table structure from the input image.
We employ TEDS~\cite{zhong2020image} as an evaluation metric.

\myparagraph{VQA for Scene Text and Document.}
For the scene text and document VQA, we include additional datasets~\cite{mishra2019ocr, gurari2018vizwiz, goyal2017making} following previous work~\cite{wang2022git}. 
We fine-tune three models with different batch ratios of the dataset: scene text VQA~\cite{singh2019towards, biten2019scene}, DocVQA~\cite{mathew2021docvqa}, and InfoVQA~\cite{mathew2022infographicvqa}.
We adopt widely employed evaluation settings in each field. 
Specifically, scene text VQA tasks are evaluated on the validation dataset~\cite{kil2022prestu}. Document VQA tasks~\cite{kim2021donut,huang2022layoutlmv3} are evaluated on the test dataset.

\myparagraph{Document Classification.}
RVL-CDIP~\cite{harley2015evaluation}, a subset of IIT-CDIP, is 400K scanned document images labeled into 16 categories. This dataset comprises 320K train images, 40K validation images, and 40K test images.

\myparagraph{Key Information Extraction.}
CORD, the Consolidated Receipt Dataset, consists of 800 train, 100 validation, and 100 test receipt images. 
We construct the target sequence the same as Donut, and the performance is reported with a TED score between generated and ground-truth JSON files. 

\subsection{Performance Evaluation and Investigation}\label{subsec:eval}
We investigate pre-training objectives and validate the effect of SCOB. 
Accordingly, we present four pre-trained models as shown in Table~\ref{tab:coverage}: 
\begin{itemize} \vspace{-0.6em}
    \item \textbf{W}$_\textnormal{OCR-read}$: a \textit{OCR-read} pre-trained model that learns a sequence composed of coordinate information and transcription in a raster scan order using Eq.~\ref{math:objective}. 
    \vspace{-0.6em}
    \item \textbf{W}$_\textnormal{OCR-read}$ w/ SCOB: a pre-trained model where SCOB is applied to \textit{OCR-read}. It is trained by Eq.~\ref{eq:total_loss} with the coordinate token of rendered data.\vspace{-0.6em} 
    \item \textbf{W}$_\textnormal{text-read}$: 
    a \textit{text-read} pre-trained model that learns transcription-only sequences in raster scan order (pseudo-label order) using Eq.~\ref{math:objective}.
     It can represent previous \textit{text-read} based methods such as Donut~\cite{kim2021donut} and Dessurt~\cite{davis2022end}.    
    \vspace{-0.6em}
    \item \textbf{W}$_\textnormal{text-read}$ w/ SCOB:
    a pre-trained model where SCOB is applied to \textit{text-read}. It is trained by Eq.~\ref{eq:total_loss} without the coordinate token. \vspace{-0.6em}
\end{itemize}
During the pre-training of \textbf{W}$_\textnormal{text-read}$, we observed several instabilities. Thus, we employed the weights of \textbf{W}$_\textnormal{text-read}$ that achieved the highest score on the validation set.
We will present more results about unstable pre-training in the supplemental file.

We also report scores of the comparison model. 
Specifically, TableFormer~\cite{nassar2022tableformer}, SPTS~\cite{peng2022spts}, and PreSTU~\cite{kil2022prestu} focused on table reconstruction, scene text OCR, and scene text VQA tasks, respectively. Additionally, Donut$_\textnormal{proto}$~\cite{kim2021donut}, Donut~\cite{kim2021donut}, and LayoutLMv3~\cite{huang2022layoutlmv3} served as foundational models for VDU tasks.
It is well known that the total capacity of GPUs is crucial for pre-training.
Unfortunately, we used only eight V100 GPUs, which is a relatively low resource compared to other methods, resulting in relatively lower performance on few benchmarks. 
However, since the main goal of our paper is to present an investigation of pre-training methods and validate the effect of SCOB, our models still provide meaningful experimental results.

\myparagraph{\textit{Text-read} vs. \textit{OCR-read}.}
We frequently observe that training \textbf{W}$_\textnormal{text-read}$ using both VDU and STU data leads to unstable learning, while \textbf{W}$_\textnormal{OCR-read}$ is trained stably. 
We suspect that the coordinate information in \textbf{W}$_\textnormal{OCR-read}$ alleviates domain conflict by guiding the text to be read explicitly even for complex scene images. Thus, \textbf{W}$_\textnormal{text-read}$ can be more vulnerable to training complex scene text images, which is represented in our experimental results.
As shown in Table~\ref{tab:coverage}, \textbf{W}$_\textnormal{OCR-read}$ considerably outperforms \textbf{W}$_\textnormal{text-read}$ on all scene text benchmarks.
On the other hand, \textbf{W}$_\textnormal{text-read}$ shows better performance on document VQA tasks. 
We think that comprehending the contextual information of the text within a well scanned or binarized image is a pivotal component of document VQA tasks. Thus, \textbf{W}$_\textnormal{text-read}$, which is trained on longer text sequences,  can be beneficial for contextual comprehension.

\myparagraph{The Effect of SCOB.}
Experimental results validate the effect of SCOB on both \textit{text-read} and \textit{OCR-read}. 
We expect SCOB to facilitate stable pre-training by bridging the complex domain gap and we confirm \textbf{W}$_\textnormal{text-read}$ w/ SCOB is trained stably.
Table~\ref{tab:coverage} also shows that SCOB considerably improves the performance on scene text OCR, VQA, and KIE benchmarks with a large margin.
In particular, it is quite notable that the improvements on TextVQA and ST-VQA are 5.1 and 4.0, respectively.
\textbf{W}$_\textnormal{OCR-read}$ w/ SCOB also notably outperforms \textbf{W}$_\textnormal{OCR-read}$ on VQA and scene text OCR benchmarks, achieving the best performance among comparisons in infoVQA. 
We would emphasize that \textbf{W}$_\textnormal{OCR-read}$ w/ SCOB is trained under the weakly supervised setting. Its required annotation is equivalent to \textbf{W}$_\textnormal{text-read}$ and \textbf{W}$_\textnormal{text-read}$ w/ SCOB, which is much lower cost than that of \textbf{W}$_\textnormal{OCR-read}$.

Large improvements are generally achieved at the OCR and VQA tasks. 
This can be because better character recognition is a prerequisite for a better understanding of document or scene text. 
Moreover, VQA is closely related to OCR because most of the answers exist in the text image. 
Significant enhancements to KIE, which involves the task of reading and organizing word boxes, arise from analogous reasons.

\myparagraph{Comparison with SoTA Methods.}
As shown in Table~\ref{tab:coverage}, the presented models achieve competitive or better performance across the VDU and STU benchmarks.
In VDU benchmarks, \textbf{W}$_\textnormal{text-read}$ achieves comparable performance to Donut and Donut$_\textnormal{proto}$~\cite{kim2021donut}. \textbf{W}$_\textnormal{text-read}$, which is considered as a re-implementation of Donut with our framework, is well reproduced considering the employed number of GPUs. 
To validate the presence of the domain gap between VDU and STU data, we fine-tune Donut on scene text VQA benchmarks where Donut is mainly pre-trained on VDU data. 
While Donut achieved a competitive advantage on the DocVQA benchmark, it only managed to secure comparable scores on ST-VQA.  Furthermore, its performance significantly dropped in the TextVQA.
We also find that read-based pre-training is effective for table reconstruction and our models achieve state-of-the-art.
In STU benchmarks, \textbf{W}$_\textnormal{OCR-read}$ w/ SCOB outperforms SPTS on all of the scene text OCR benchmarks and achieves better performance than PreSTU on TextVQA.

\myparagraph{Comparison with Larger Models.}
Our methodologies are compared with more recent approaches such as UDOP~\cite{tang2023unifying}, PaLI~\cite{chen2023pali}, and GIT2~\cite{wang2022git} as presented in Table~\ref{tab:coverage}. 
UDOP and PaLI-17B exhibit superior performance across various benchmark criteria. 
These results show the potential of leveraging OCR to enhance the performance of generative models, albeit with concomitant resource implications.
Based on PaLI-3B and PaLI-17B results, increasing the model size can also be a significant factor in the performance. 
Furthermore, it is noteworthy that PaLI employs a form of \textit{text-read} task as a pre-training method. In this context, we believe that the integration of our SCOB approach holds the potential to further amplify the performance of PaLI.

{
\setlength{\tabcolsep}{4pt}
\renewcommand{\arraystretch}{1.3} 
\begin{table}[t]
\begin{center}

    \resizebox{0.87\linewidth}{!}{%
        \begin{tabular}{@{}lcccc@{}} \toprule
            Method & \#Params &\makecell[c]{CORD~\cite{park2019cord}\\(Acc)} & \makecell[c]{RVL-CDIP~\cite{harley2015evaluation}\\(Acc)} & \makecell[c]{DocVQA~\cite{mathew2021docvqa}\\(ANLS)} \\
            \midrule
            BERT~\cite{kenton2019bert}        &110M + $\alpha$& 65.5  & 89.8  & 63.6 \\
            LayoutLM~\cite{xu2020layoutlm}    &113M + $\alpha$& 81.3  & 94.4  & 69.8 \\
            LayoutLMv2~\cite{xu2021layoutlmv2}  &200M + $\alpha$& 82.4  & 95.3  & 78.1 \\
            LayoutLMv3~\cite{huang2022layoutlmv3} &133M + $\alpha$ & 84.4 & \textbf{95.4}  & \textbf{78.8} \\
            Dessurt~\cite{davis2022end} &127M & - & 93.6 & 63.2 \\
            Donut$_\textnormal{proto}$~\cite{kim2021donut}       &143M & 85.4  & 94.5  & 47.1 \\
            Donut~\cite{kim2021donut}       &143M& \textbf{90.9}  & 95.3  & 67.5 \\
            \midrule
            W$_\textnormal{OCR-read}$  &202M& 88.2 & 94.2 & 56.1 \\ 
            W$_\textnormal{OCR-read}$ w/ SCOB &202M& 88.5 & 94.6 & 60.5 \\ 
            W$_\textnormal{text-read}$  &202M& 84.4 & 94.4 & 57.0 \\ 
            W$_\textnormal{text-read}$ w/ SCOB &202M& 87.4 & 94.3 & 59.6 \\ 
            
            \bottomrule
        \end{tabular}}
    \caption{The public benchmark on CORD~\cite{park2019cord}, RVL-CDIP~\cite{harley2015evaluation}, and DocVQA~\cite{mathew2021docvqa}. 
    $\alpha$ is represented for the requirement of an additional OCR model.
    }
    \label{tab:vdus}
     \end{center}\vspace{-2.0em}
\end{table}  
}

\setlength{\tabcolsep}{4pt}
\renewcommand{\arraystretch}{1.3} 
\begin{table}[t]
\begin{center}

    \resizebox{0.7\linewidth}{!}{%
        \begin{tabular}{@{}lccc@{}} \toprule
            Method & \#Param &   \makecell[c]{TextVQA~\cite{singh2019towards}\\ (Acc.)}  & \makecell[c]{ST-VQA~\cite{biten2019scene}\\(ANLS)} \\ \midrule
            SA-M4C ~\cite{kant2020spatially}  & 93M &  45.4 & 51.2 \\ %
            TAP~\cite{yang2021tap} & 160M &  54.7 & 59.8 \\
            GIT$_{Large}$~\cite{wang2022git} & 347M &   37.5 & 44.6 \\ 
            PreSTU~\cite{kil2022prestu} & 278M &  54.5 & 62.6 \\ \midrule

            W$_\textnormal{OCR-read}$  &202M& 55.4 & \textbf{62.9} \\
            W$_\textnormal{OCR-read}$ w/ SCOB &202M & \textbf{56.2} & 62.6 \\
            W$_\textnormal{text-read}$  &202M& 49.3 & 57.2 \\
            W$_\textnormal{text-read}$ w/ SCOB &202M & 54.4 & 61.2 \\            
            \midrule
            \midrule
            Flamingo~\cite{alayrac2022flamingo} & 80B &   57.1 & - \\
            GIT~\cite{wang2022git} & 681M &  59.9 & 69.1 \\
            LaTr~\cite{biten2022latr} + Rosetta-en & n/a & 48.4 & - \\
            LaTr~\cite{biten2022latr} + Amazon-OCR & n/a & 59.5 & 67.5 \\
            \bottomrule
        \end{tabular}}
    \caption{The public benchmark on TextVQA~\cite{singh2019towards} and ST-VQA~\cite{biten2019scene} for scene text VQA. LaTr requires OCR results as input. We report two results depending on the employed OCR models.
    The best performance among similar-sized models (\#Param less than 400M) is represented in \textbf{bold}.}
    \label{tab:scenetext_vqa}
     \end{center}\vspace{-1.5em}
\end{table}

\subsection{Detailed Performance Comparison}
In this subsection, we compare our models with more diverse methods.
Table~\ref{tab:vdus} shows that our models present the second-best performance on CORD and relatively lower performance on DocVQA.
LayoutLMv2~\cite{xu2021layoutlmv2} and LayoutLMv3~\cite{huang2022layoutlmv3} have notable performance on DocVQA. 
Unlike Donut, Dessurt, and our models, LayoutLMs take both image and text (OCR) modalities as input, which additionally requires OCR results as a pre-process.
Accordingly, LayoutLMv3 (1.8 sec/img) takes longer inference time than our models (1.1 sec/img) due to acquiring OCR results. We measure the inference time with a V100 GPU on CORD dataset~\cite{park2019cord}. 
Kim~\etal~\cite{kim2021donut} also reported that Donut is 2 times faster than LayoutLMv2. 

As illustrated in Table~\ref{tab:scenetext_vqa}, our models show comparable performance on scene text VQA tasks. Specifically, our model achieves the best performance among the similar-sized models. 
Surprisingly, W$_\textnormal{OCR-read}$ w/ SCOB achieves comparable performance to Flamingo, which has extremely large parameters. 
Since our models are also scalable, like GIT, we expect the enlarged models to reach the performance of GIT and Flamingo. 
We discuss more comparisons on other tasks such as table reconstruction, layout analysis, and scene text OCR, in a supplemental file.

{
\setlength{\tabcolsep}{4pt}
\renewcommand{\arraystretch}{1.3} 
\begin{table}[t]
\begin{center}

\resizebox{0.95\linewidth}{!}{%
\begin{tabular}{@{}lcc|cc@{}} \toprule
Components & KIE~\cite{park2019cord} & DocVQA~\cite{mathew2021docvqa} & OCR~\cite{karatzas2013icdar, karatzas2015icdar, ch2017total} & Scene VQA~\cite{singh2019towards,biten2019scene} \\\midrule
A. \textit{OCR-read}              & 88.2                 & 55.1                       & 81.3       & 57.3    \\
B. A w/ SupCon              & 88.0                 & 50.0                    & 82.2          &  56.8  \\
C. A w/ rendering              & 87.7                 & 47.8                       & 82.0      & 54.3   \\
D. A w/ SCOB    & \textbf{88.5}                 & \textbf{55.5}                    & \textbf{83.0}       & 59.4 \\
E. D w/ full annotation & 86.8                 & 55.1                    & 82.6       & \textbf{59.6} \\ \bottomrule
\end{tabular}
}
\caption{Ablation study on the proposed components. E denotes that the model is trained by SCOB with full annotations of both rendered and real images. We report the performance averaged on scene text OCR and scene text VQA.}
\label{tab:ablations}
\end{center}\vspace{-2.0em}
\end{table}
}

\section{Analysis} \vspace{-0.5em}
\myparagraph{Ablation Study.}
We conduct an ablation study on proposed components: (A) \textit{OCR-read}, (B) character-wise SupCon, (C) online text rendering, (D) SCOB, and (E) SCOB with full annotations of rendered and real images.
We pre-train each model with different components and fine-tune each model on several downstream tasks.
As shown in Table~\ref{tab:ablations}, B and C improve the performance of OCR but degrades that of the other downstream tasks. 
We think the reasons are as follows:
i) For the case of SupCon, a naive augmentation used in previous works~\cite{grill2020bootstrap, he2020momentum, chen2020simple, chen2020improved} would not be beneficial to other downstream tasks (compare A vs. B).
ii) For the case of online text rendering, it is trained only with half of the real data because half of the batch is charged with rendered images (compare A vs. C).
iii) For the case of SCOB, the renderer plays a critical role as a fitted augmentation of SupCon by providing strong variance to positive samples. Also, character-wise SupCon bridges all synthetic, document, and scene domains by enforcing the same characters close together, which presents the synergistic effect.
Comparing D and E, SCOB using full supervision could not significantly improve the performance, which shows SCOB is successfully pre-trained with weak supervision.
We conduct an ablation study with a down-scaled setting for efficiency. More details will be provided in the supplementary material. 

\begin{figure}[t]
\begin{center}
   \includegraphics[width=.8\linewidth]{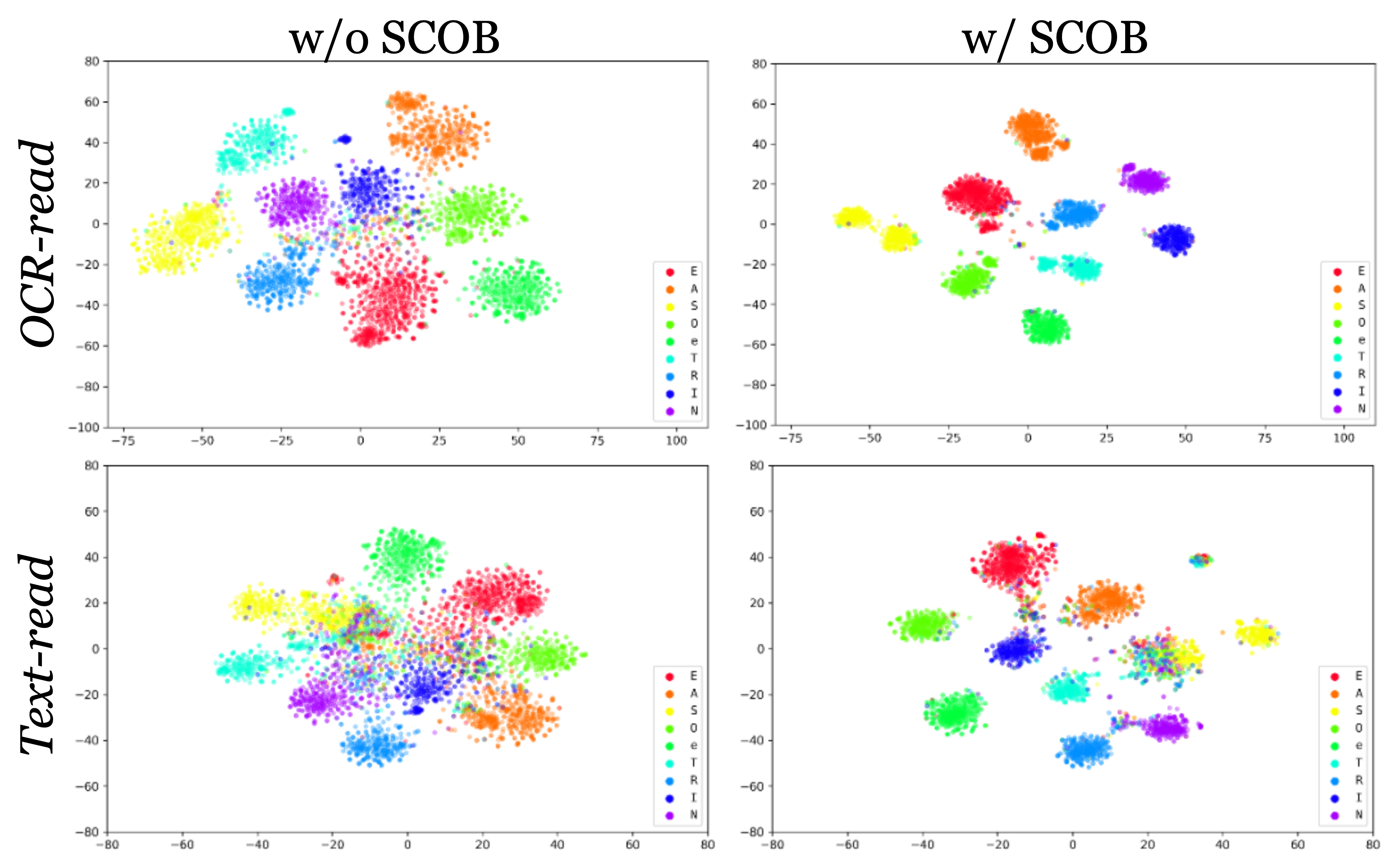}
\end{center}\vspace{-1.0em}
  \caption{The t-SNE~\cite{tsne2008} visualization for \textit{OCR-read}, \textit{text-read}, and their respective SCOB applications. Note that different colors denote each class (character) and the nine most predicted characters are displayed. \textit{Data}: ICDAR2015~\cite{karatzas2015icdar} test set}  
\label{fig:TSNE}\vspace{-1.2em}
\end{figure}

\myparagraph{Qualitative Analysis.}
In Figure~\ref{fig:TSNE}, we visualize representations extracted from the final layer of the decoder using t-SNE~\cite{tsne2008} by mapping high-dimensional features into low-dimensional space through KL-divergence. Note that our SCOB clusters embeddings more discriminatively than other methods. We will provide more various figures in the supplementary material.

\section{Conclusion}
\label{sec:conclusion}
This paper investigates an effective pre-training on a total of eleven text-related tasks in the document and scene text domains. Our proposed SCOB is a new pre-training method for universal text understanding that leverages a character-wise supervised contrastive loss with online text rendering, enhancing the stability of training and reducing annotation costs. Experimental results on various visual text-related tasks validate that our SCOB is broadly applicable to read-based pre-training methods and improves performance.

\section{Acknowledgements}
We greatly appreciate Teakgyu Hong and Han-Cheol Cho for their help with the initial project setup.

{\small
\bibliographystyle{ieee_fullname}
\bibliography{egbib}
}

\newpage
\appendix

\section*{Content}
This appendix provides details of pre-training experiments (Section~\ref{sec:ap_intro}), detailed settings of fine-tuning (Section~\ref{sec:finetune}), and more diverse samples of t-SNE~\cite{tsne2008} (Section~\ref{sec:visual}).

\section{Pre-trainings}
\label{sec:ap_intro}

\subsection{Details of Pre-training} 
In Pix2Seq~\cite{chen2021pix2seq}, sequence augmentation was applied to add a noise token to the sequence. On the other hand, we do not apply sequence augmentation to our model because we could not empirically observe the performance improvement with noise tokens. 
The learning rate is set to 1e-4 with a cosine decay scheduler~\cite{CosineLR} where a warm-up step is 100K steps.
For the data augmentation, we adopt the widely used augmentation methods for OCR as follows: random crop following CRAFT~\cite{baek2019character}, random rotation, random resize options (e.g., bilinear, nearest neighbor, bicubic, and Lanczos interpolation), and photometric distortion. 

\begin{figure*}[b]
\begin{center}
   \includegraphics[width=0.9\textwidth]{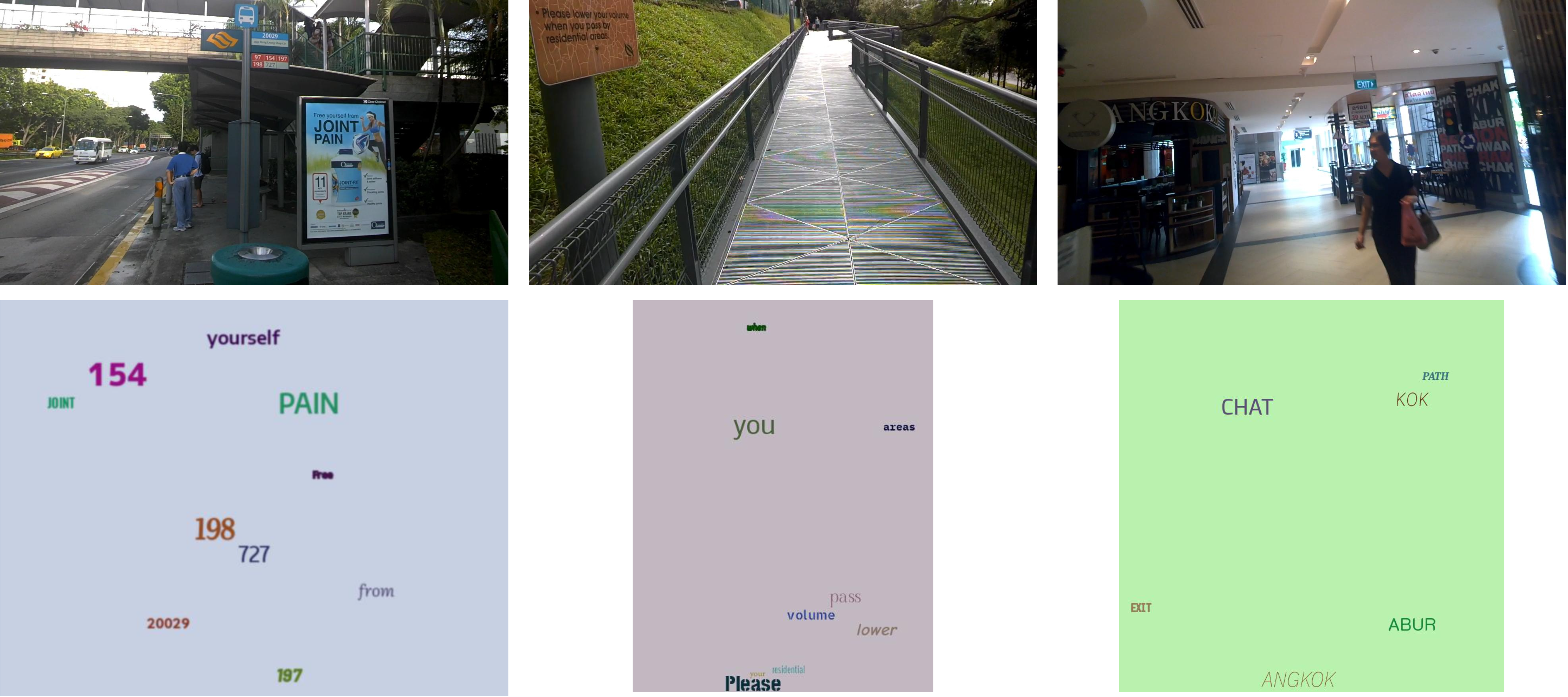}
\end{center}
   \caption{
   The visualization of images generated in a one-to-one correspondence to the ICDAR2015 test set using the proposed online renderer.
   } 
   \label{subfig:otor_exam}
\end{figure*}

The detailed settings of the online text renderer are as follows. The range of resolution is from 400 to 768, the range of background RGB values is from 151 to 255, and the Gaussian blur radius value is 0 to 1.8 with a probability of 0.2. Figure~\ref{subfig:otor_exam} provides image samples generated in a one-to-one correspondence to the ICDAR2015~\cite{karatzas2015icdar} test set using the renderer.

\subsection{Instability of \textit{Text-read}} 
In this section, we present experimental results about the instability of \textit{text-read} when the model is pre-trained with both VDU and STU data.    
As significant performance differences are observed in the early stages of training, we conduct a precise experiment.
Specifically, we pre-train the model using IIT-CDIP and scene text datasets for 200K steps with a batch size of 16.
For fine-tuning OCR, we employ scene text datasets used in the main paper and fine-tune the models for 50K training steps with 8 batch size that consists of $1920\times1920$ resolution images. 

\begin{figure}[t]
\begin{center}
    \includegraphics[width=1.\linewidth]{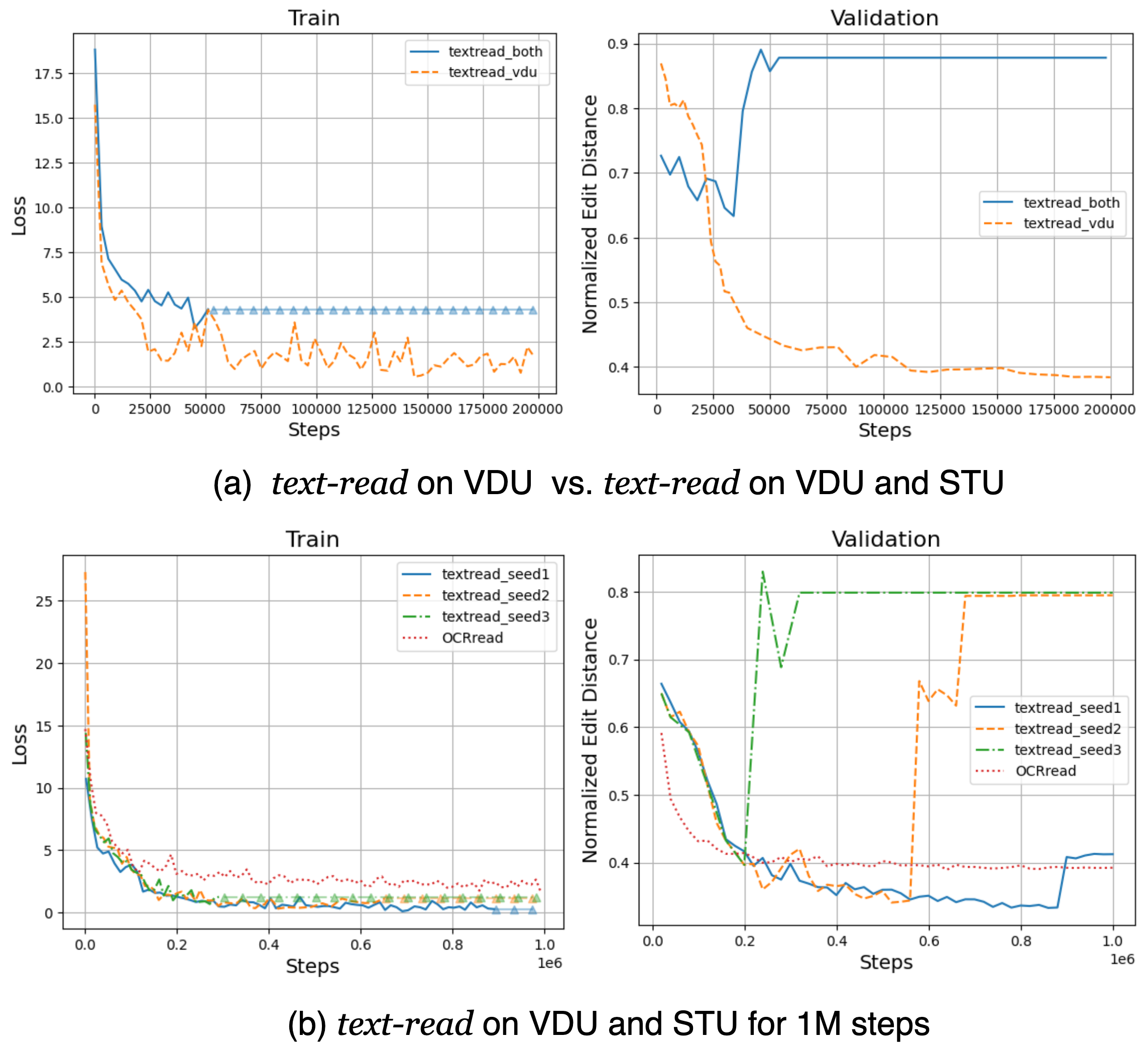}\vspace{-0.5em}
\end{center}
   \caption{
   Graphs of training loss and validation score during \textit{text-read} pre-training.   
   (a) shows instability of \textit{text-read} using VDU and STU together. (b) also shows instability of \textit{text-read} using VDU and STU together for 1M steps compared to \textit{OCR-read}.
   Validation score is normalized edit distance. Upper triangle in loss graph means NaN (not a number).}
   \label{subfig:domain_conflict}
\end{figure}

\begin{table*}[]
\begin{center}

\resizebox{0.8\linewidth}{!}{%
\begin{tabular}{@{}lcc|ccccc@{}} \toprule
\multirow{2}{*}{Method} & KIE  & Document VQA & \multicolumn{3}{c}{OCR} & \multicolumn{2}{c}{Scene Text VQA} \\ \cmidrule{2-8}
& CORD~\cite{park2019cord} & DocVQA~\cite{mathew2021docvqa}       & IC13~\cite{karatzas2013icdar} & IC15~\cite{karatzas2015icdar} & TotalText~\cite{ch2017total} & TextVQA~\cite{singh2019towards}          & ST-VQA~\cite{biten2019scene}  \\
\midrule
\textit{Text-read} on both domains  & 80.7                 & 45.7                       & 66.7 & 20.3 & 43.4       & 42.1           & 52.2         \\
\textit{Text-read} on VDU              & 85.3                 & 53.1                    & 77.1   & 35.8 & 56.1           & 42.8           & 53.2         \\
\bottomrule
\end{tabular}
}
\caption{The performance evaluation for \textit{text-read} according to pre-training domains. We report the end-to-end recognition F-measure scores on IC13, IC15, and TotalText evaluated with strong, strong, and full lexicons, respectively}
\label{suptab:vs}
\end{center}
\end{table*}

Figure~\ref{subfig:domain_conflict} shows the convergence of two distinct reading approaches: \textit{text-read} and \textit{OCR-read}. 
Specifically, Figure~\ref{subfig:domain_conflict} (a) shows that the loss and validation score converge stably when only VDU data is used as a training set.
However, the model fails to generate appropriate sequences during validation and experiences NaN (not a number) loss when both VDU and STU datasets are used for training. 
Figure~\ref{subfig:domain_conflict} (b) depicts the \textit{text-read} pre-training process for 1M steps, which has failed across various seed values. 
On the other hand, \textit{OCR-read} displays more stable convergence patterns despite its loss and validation score converging at a slightly higher level due to sensitivity issues such as coordinate token output.
The sole difference between \textit{OCR-read} and \textit{text-read} lies in the presence or absence of coordinate prediction. Also, the distinction between VDU and STU domains is based on whether text is present in the natural scene background.
We posit that a domain gap exists between VDU and STU, contributing to the instability of \textit{text-read}.

To evaluate the performance of the pre-trained weights on downstream tasks, we fine-tuned \textit{text-read} on four downstream tasks using pre-trained weights for 200K steps ((a) of Figure~\ref{subfig:domain_conflict}). Table~\ref{suptab:vs} indicates that \textit{text-read} pre-training on both domains fails to transfer learning on downstream tasks. The performance of \textit{text-read} using both domains is relatively lower than that of \textit{text-read} using only the VDU domain. This result further supports that \textit{text-read} is unstable in both domains, deteriorating the achievement of a reliable pre-training.


\subsection{Details of Ablation Study} 
\begin{table*}[]
\begin{center}
\resizebox{0.8\linewidth}{!}{%
\begin{tabular}{@{}lcc|ccccc@{}} \toprule
\multirow{2}{*}{Method} & KIE  & Document VQA & \multicolumn{3}{c}{OCR} & \multicolumn{2}{c}{Scene Text VQA} \\ \cmidrule{2-8}
& CORD~\cite{park2019cord} & DocVQA~\cite{mathew2021docvqa}       & IC13~\cite{karatzas2013icdar} & IC15~\cite{karatzas2015icdar} & TotalText~\cite{ch2017total} & TextVQA~\cite{singh2019towards}          & ST-VQA~\cite{biten2019scene}  \\
\midrule
A. W$_{\textnormal{OCR-read}}$              & 88.2                 & 55.1                       & 94.3 & 74.2 &75.5       & 55.4           & 59.2         \\
B. A w/ SupCon               & 87.7                 & 50.0                    & 93.4   & 76.5 & \textbf{76.7}           & 53.0           & 60.6         \\
C. A w/ rendering                 & 88.0                 & 47.8                       & 94.1 & 76.4 & 75.4       & 50.8           & 57.7         \\
D. A w/ SCOB        & \textbf{88.5}                 & \textbf{55.5}                    & \textbf{95.0} & 77.6 & 76.5       & \textbf{56.2}           & 62.6         \\

E. D w/ full annotation& 86.8                 & 55.1                    & 94.7 & \textbf{77.7} & 75.5       & 56.1           & \textbf{63.1}      \\ \bottomrule

\end{tabular}
}
\caption{Ablation study on the proposed components. For the evaluation of OCR, we report the end-to-end recognition F-measure scores on IC13, IC15, and TotalText evaluated with strong, strong, and full lexicons, respectively.}
\label{suptab:ablations_big}
\end{center}
\end{table*}
As presented in Table 4 of the main paper, we conduct an ablation study with five models. All models are pre-trained for 1M steps. Table~\ref{suptab:ablations_big} provides detailed performance of benchmarks.
Details are as follows:
\begin{itemize}
\item A: W$_{\textnormal{OCR-read}}$. \vspace{-0.5em}
\item B: We introduce augmentation for generating multiview images. The augmentation types are random rotation, image resize, color jittering, converting into the gray channel, and Gaussian blur. Geometric augmentations such as random rotation and image resize are applied with low intensity. The model is trained via Eq. 1 and Eq. 2. \vspace{-0.5em}
\item C: The real and rendered data are the input data of the model. The model is trained via Eq. 3, and the renderer is the same as SCOB. \vspace{-0.5em} 
\item D: W$_{\textnormal{OCR-read}}$ with SCOB. \vspace{-0.5em}
\item E: The model is trained under the fully supervised learning. Thus, the coordinate information of real data is employed for pre-training.
\end{itemize}
We fine-tune 4 downstream tasks such as KIE, document VQA, OCR, and scene text VQA. 
For the tasks of VQA, we evaluate the validation set, because multiple submissions of the test dataset on the leaderboard can be regarded as cheating. 
We fine-tuned models for 50K steps using 16 batch size on VQAs for efficiency.
For the OCR tasks, we employ intermediate fine-tuned models, which are evaluated on the test dataset.

\begin{figure}[t]
\begin{center}
    \includegraphics[width=1.0\linewidth]{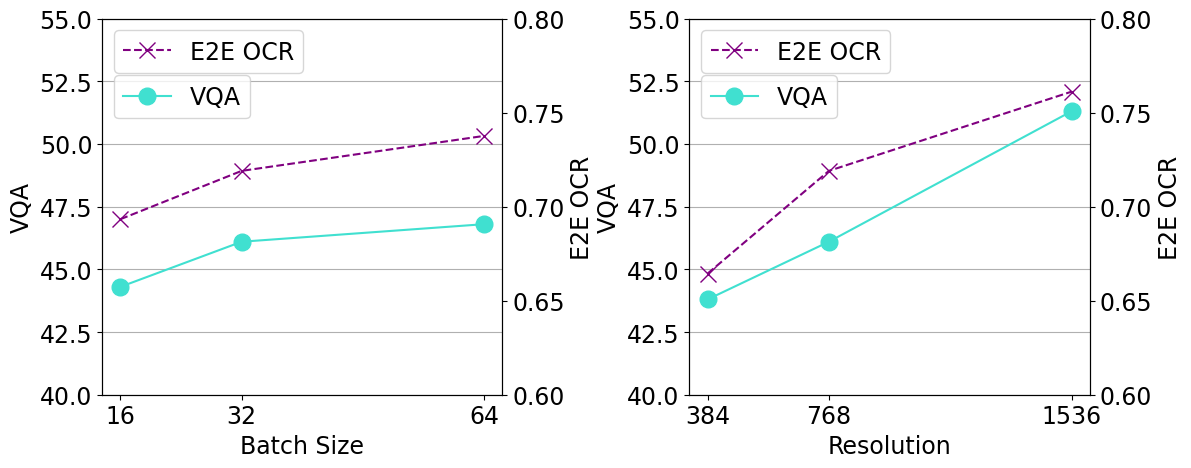}\vspace{-0.5em}
\end{center}
   \caption{The results for VQA (average scores of DocVQA, TextVQA, STVQA) and E2E OCR (average F1 scores of IC13, IC15, TotalText) across batch sizes (16, 32, 64) and resolutions (384, 768, 1536). A fixed resolution of 768 and batch size of 32 were used respectively. The experiment involved 200K steps of \textit{OCR-read} pre-training and 50K of fine-tuning.}
   \label{subfig:batchsize_resolution}
\end{figure}

\subsection{Performance Trends with Batch Size and Resolution}
We present the experimental results of the effect of batch size and resolution changes on downstream task performance under the \textit{OCR-read} pre-training setting in Figure~\ref{subfig:batchsize_resolution}. The results confirm an enhancement in performance as both batch size and resolution increase. We identified a performance gap between SCOB and the current SoTA. However, in this paper, we delve into the composite domain of VDU and STU to compare the performance of pre-training tasks, verifying the superiority of SCOB under identical pre-training environments. Based on these results, we expect that there is latent potential within the model, likely to be uncovered through the careful tuning of factors such as scale, learning schedule, and resolution.

\section{Fine-tunings}
\label{sec:finetune}

\begin{table*}[t!]
\begin{center}
\resizebox{1.0\textwidth}{!}{%
\begin{tabular}{@{}lcccccccc@{}} \toprule
Task & Steps & Batch Size & LR & GC & DPL & MSL & Resolution & Batch Ratio of Fine-tuning Dataset\\ \midrule
\midrule
Table Reconstruction* & 400K & 16 &  5e-5 & 1.0 & 6 & 3072 & 768 &  \textbf{1.0} PubTabNet~\cite{zhong2020image}  \\
\midrule
KIE & 200K & 16 & 3e-5 & 1.0 & 0 & 512 & 1920 &  \textbf{1.0} CORD~\cite{park2019cord}  \\
\midrule
Document Classification & 1M & 16 & 2e-5 & 1.0 & 0 & 8 & 1920 &   \textbf{1.0} RVL-CDIP~\cite{harley2015evaluation} \\
\midrule
Document VQA & 100K & 16 & 3e-5& 0.25 & 0 & 512 & 1920 & \makecell[c]{\textbf{0.65} DocVQA~\cite{mathew2021docvqa}, \textbf{0.07} TextVQA~\cite{singh2019towards} ,  \textbf{0.07} ST-VQA~\cite{biten2019scene}, \\  \textbf{0.07} OCRVQA~\cite{mishra2019ocr},  \textbf{0.07} VizWiz~\cite{gurari2018vizwiz},  \textbf{0.07} VQAv2~\cite{goyal2017making}}\\
\midrule
Infographics VQA & 100K & 16 & 3e-5 & 0.25 & 0 & 512 & 1920 & \makecell[c]{ \textbf{0.58} InfoVQA~\cite{mathew2022infographicvqa}, \textbf{0.07} DocVQA~\cite{mathew2021docvqa},  \textbf{0.07} TextVQA~\cite{singh2019towards}, \\ \textbf{0.07} ST-VQA~\cite{biten2019scene},  \textbf{0.07} OCRVQA~\cite{mishra2019ocr}, \textbf{0.07} VizWiz~\cite{gurari2018vizwiz},   \textbf{0.07} VQAv2~\cite{goyal2017making}}\\ 
\midrule
Layout Analysis & 100K & 16 & 2e-4 & 1.0 & 0 & 512 & 1536 &  \textbf{1.0} PubLayNet~\cite{zhong2019publaynet}\\
\midrule
Scene Text VQA & 100K & 16 & 3e-5 & 0.25 & 0 & 512 & 1920  & \makecell[c]{ \textbf{0.2} DocVQA~\cite{mathew2021docvqa},  \textbf{0.2} TextVQA~\cite{singh2019towards},  \textbf{0.2} ST-VQA~\cite{biten2019scene}, \\ \textbf{0.2} OCRVQA~\cite{mishra2019ocr},  \textbf{0.1} VizWiz~\cite{gurari2018vizwiz}, \textbf{0.1} VQAv2~\cite{goyal2017making}} \\
\midrule
Scene Text OCR* & 100K & 16 & 5e-5 & 1.0 & 0 & 512 & 2560 & \makecell[c]{\textbf{0.3} HireText~\cite{long2022towards} , \textbf{0.3} TextOCR~\cite{singh2021textocr}, \textbf{0.09} TotalText~\cite{ch2017total}, \\ \textbf{0.09} OpenImagesv6~\cite{OpenImages2}, \textbf{0.01} IC13~\cite{karatzas2013icdar}, \textbf{0.01} IC15~\cite{karatzas2015icdar}} \\
\bottomrule
\end{tabular}
}
\caption{Detailed settings of downstream tasks. For the batch ratio, we represent the cell as `\textit{\textbf{batch ratio of dataset} dataset}'. 
\textit{Abbr.} \textit{LR}: learning rate, \textit{GC}: gradient clipping, \textit{DPL}: decoder prune layer, \textit{MSL}: maximum sequence length of decoder. * denotes that not using intermediate training.
}\label{suptab:details}
\end{center}
\end{table*}

\subsection{Details of Fine-tuning} 
In the fine-tuning stage, we use $1920\times1920$ image resolution to recognize text well, and intermediate training follows it. Moreover, we empirically confirmed that auxiliary loss~\cite{carion2020end} accelerates the convergence of the recognition performance of OCR. Thus we apply the auxiliary loss to all layers of the decoder in the fine-tuning. 

Table~\ref{suptab:details} provides our detailed settings for fine-tuning. 
We fine-tune all models using Adam optimizer with a cosine decay scheduler where the warm-up step is set to 10\% of training steps.
We have not widely explored the hyperparameters, such as the batch ratio of the dataset, gradient clipping value, and learning rate. 
We believe more hyperparameter exploration will result in better performance. For the input image resolution, we determine the resolution considering the original image size and comparisons of the input size.

\begin{figure*}[t]
\begin{center}
   \includegraphics[width=0.9\textwidth]{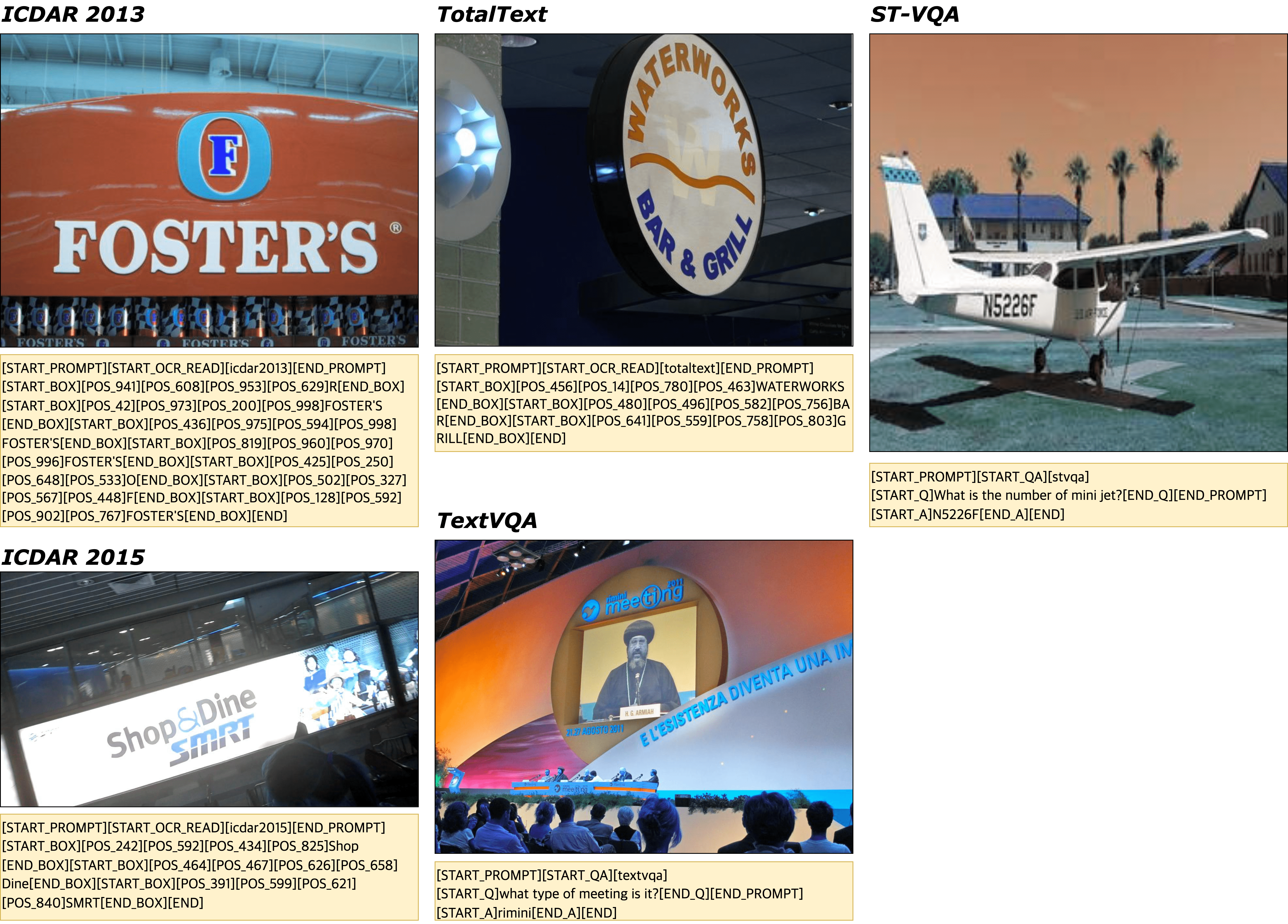}
\end{center}
   \caption{
   Examples of STU tasks. We provide a sample image for each dataset and its ground truth sequence.
   } 
   \label{subfig:stu_samples}
\end{figure*}

\begin{figure*}[t]
\begin{center}
   \includegraphics[width=0.9\textwidth]{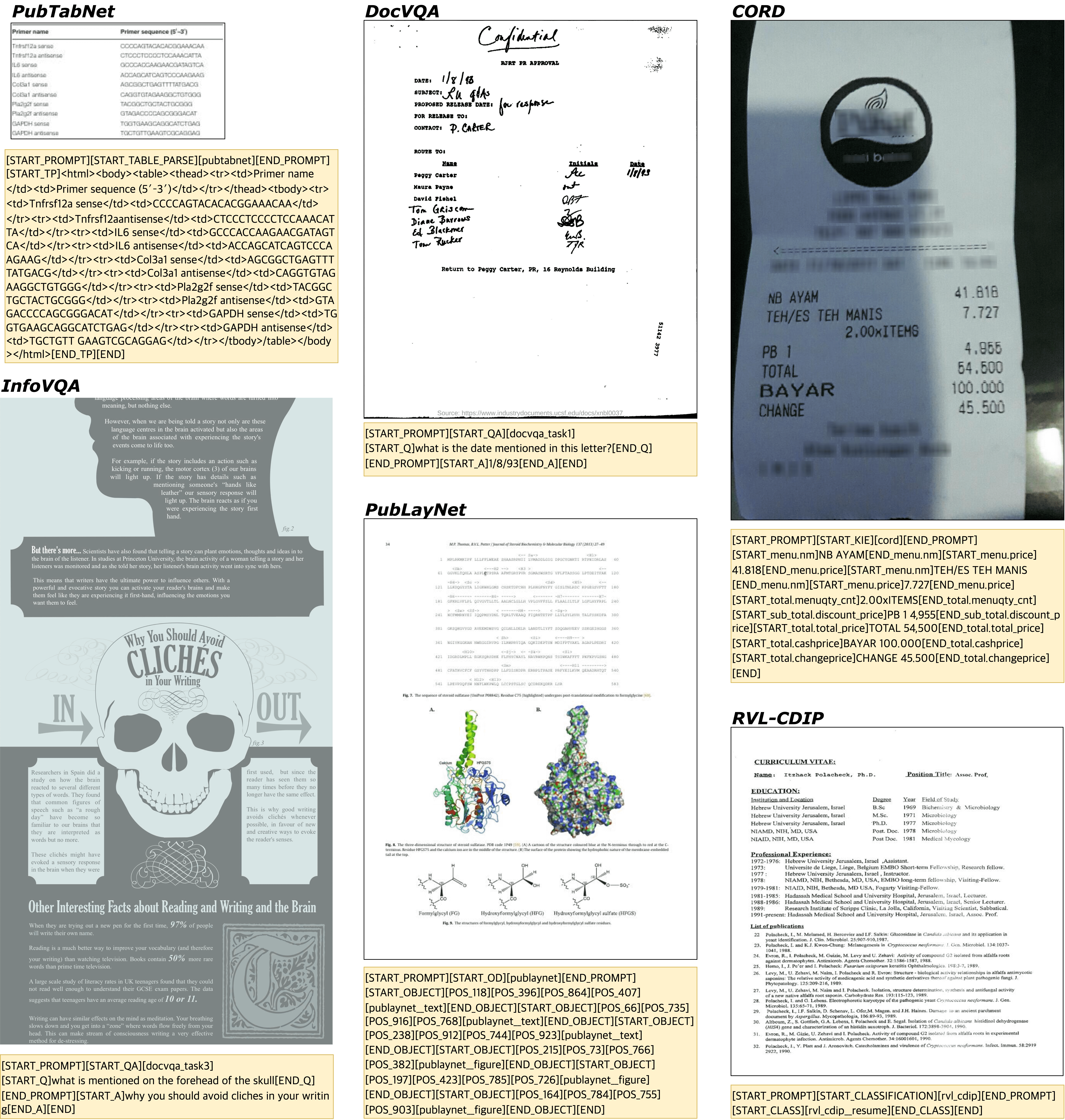}
\end{center}
   \caption{
   Examples of VDU tasks. We provide a sample image for each dataset and its ground truth sequence. Note that a sample image of InfoVQA dataset is too long vertically, thus we crop the answer part of the question and report that.
   } 
   \label{subfig:vdu_samples}
\end{figure*}

Here, we supplement more details of specific downstream tasks.
To help understand downstream tasks, we provide the examples of image and its ground truth sequence of our model in Figures~\ref{subfig:stu_samples} and~\ref{subfig:vdu_samples}. 

\myparagraph{Table Reconstruction.} In table reconstruction, the model should decode table structures and contents in the cell.
Thus, the maximum sequence length of the decoder is set to 3072, which makes pruning 6 Transformer layers in the decoder to reserve batch size 16. 
In table reconstruction, the input can be categorized as image and PDF.

\myparagraph{Layout Analysis.}
PubLayNet~\cite{zhong2019publaynet} is a dataset annotated with a bounding box format and 5 document layout categories: text, title, list, figure, and table. This includes 335,703 training images and 11,245 validation images. We fine-tune the train set and evaluate the validation set, following LayoutLMv3~\cite{huang2022layoutlmv3}.

\myparagraph{Scene Text OCR.} 
To evaluate the performance of our OCR model, we assess its ability to recognize text in different benchmarks, including ICDAR2013~\cite{karatzas2013icdar}, ICDAR2015~\cite{karatzas2015icdar}, and TotalText~\cite{ch2017total}. However, since each dataset has a distinct format for coordinate annotation, we conduct short annotation fine-tuning specifically for IC15 and TotalText to mitigate any discrepancies. Table~\ref{suptab:ocr_details} presents the performance of our model on IC13, IC15, and TotalText with additional lexicon conditions. Note that the OCR performance reported in Table~\ref{suptab:ablations_big} is obtained before annotation fine-tuning. 

{
\setlength{\tabcolsep}{4pt}
\renewcommand{\arraystretch}{1.3} 
\begin{table}[t]
\begin{center}
    \resizebox{0.95\linewidth}{!}{%
        \begin{tabular}{@{}lcccc|cccc|cc@{}} \toprule
            \multirow{2}{*}{Method} & \multicolumn{4}{c}{IC13 End-to-End} & \multicolumn{4}{c}{IC15 End-to-End} & \multicolumn{2}{c}{Total-Text}  \\ \cmidrule{2-11}
            &  \textit{N} & \textit{G} & \textit{W} & \textit{S} & \textit{N} & \textit{G} & \textit{W} & \textit{S} & \textit{N} & \textit{F} \\  
            \midrule
            W$_\textnormal{OCR-read}$ &90.4 & 91.8 & 93.7 & 94.3 &66.4  & 69.4& 73.3 & 75.7 & 72.6 & 78.9\\ 
            W$_\textnormal{OCR-read}$ w/ SCOB &\textbf{92.2}& \textbf{93.1} & \textbf{94.6} & \textbf{95.0} &\textbf{71.9} & \textbf{74.2}& \textbf{76.6} &\textbf{78.8} & \textbf{75.1} & \textbf{79.8} \\
            W$_\textnormal{text-read}$ &87.7 &89.2   &91.1   &91.8   &55.6   &58.7   &61.3   &63.3   &69.1   &75.1\\ 
            W$_\textnormal{text-read}$ w/ SCOB &88.6    &90.0   &91.7   &   92.1    &59.3   &61.1   &63.1   &65.5   &73.1   &78.2\\ 
            
            \bottomrule
        \end{tabular}}
     \end{center}\vspace{-1.em}
    \caption{The end-to-end recognition results on ICDAR2013, ICDAR2015, and Total-Text. \textit{Abbr.}~~\textit{N, G, W, S, F}: none, generic, weak, strong, and full lexicons, respectively.} 
    \label{suptab:ocr_details}
\end{table}
}

\subsection{Comparison on Table Reconstruction}

{
\setlength{\tabcolsep}{4pt}
\renewcommand{\arraystretch}{1.3} 
\begin{table}[t]
\begin{center}
    \resizebox{.7\linewidth}{!}{%
        \begin{tabular}{@{}llccc@{}} \toprule
            \multirow{2}{*}{Input} & \multirow{2}{*}{Method} & \multicolumn{3}{c}{TEDS}  \\ \cmidrule{3-5}
            & & Simple & Complex & All \\ \midrule
            \multirow{3}{*}{PDF} & Tabula & 78.0 & 57.8 & 67.9 \\
            & Acrobat Pro &68.9 &61.8 &65.3 \\
            & TableFormer~\cite{nassar2022tableformer} &95.4 &90.1 &93.6 \\ \midrule
            \multirow{7}{*}{Image} & Acrobat Pro &53.8 &53.5 &53.7 \\
            & WYGIWYS~\cite{deng2017image} & 81.7 & 75.5 & 78.6 \\
            & EDD~\cite{zhong2020image} &91.2 &85.4 &88.3 \\ \cmidrule{2-5}
            & W$_\textnormal{OCR-read}$ & 97.7 & 94.2 & 96.0 \\
            & W$_\textnormal{OCR-read}$ w/ SCOB & 97.5 & 94.1 & 95.9\\
            & W$_\textnormal{text-read}$ &\textbf{97.9} & \textbf{94.5} & \textbf{96.2} \\ 
            & W$_\textnormal{text-read}$ w/ SCOB & 97.6 & 94.1 & 96.0 \\
            \bottomrule
        \end{tabular}}
    \caption{The public benchmark on PubTabNet~\cite{zhong2020image} for table reconstruction including content in the cell. The best performance is represented in \textbf{bold} including the rest of the tables.}
    \label{tab:table_recon}
     \end{center}\vspace{-1.5em}
\end{table}
}

Taking the image input is more challenging than PDF because the contents in the cell should also be decoded. 
As can be seen in Table~\ref{tab:table_recon}, the performance of Acrobat Pro on PDF has much higher than that on the image input.
Our models substantially outperform previous methods despite using the image input and can be a strong baseline for the generation model.

\subsection{Comparison on Layout Analysis} 

{
\setlength{\tabcolsep}{4pt}
\renewcommand{\arraystretch}{1.3} 
\begin{table}[t]
\begin{center}
    \resizebox{0.95\linewidth}{!}{%
        \begin{tabular}{lcccccccc} \toprule
            \multirow{2}{*}{Method} &\multirow{2}{*}{\#Param}& \multirow{2}{*}{\textit{G}} & \multicolumn{5}{c}{Category} & \multirow{2}{*}{mAP}  \\ \cmidrule{4-8}
            &  &  & Text & Title & List & Table & Figure & \\ \midrule
            PubLayNet~\cite{zhong2019publaynet} &-&& 91.6 & 84.0 & 88.6 & 96.0 & 94.9 & 91.0 \\
            LayoutLMv3~\cite{huang2022layoutlmv3} &133M &&94.5 &90.6 &95.5 &97.9 &97.0 & \textbf{95.1} \\
            UDoc~\cite{gu2022unified} &272M  & &93.9 &88.5 &93.7 &97.3 &96.4 &93.9 \\
            DiT~\cite{li2022dit} &304M & & 94.4 &89.3 &96.0 & 97.8 & 97.2 & 94.9 \\ \midrule
            W$_\textnormal{OCR-read}$ &202M &\checkmark & 93.1 & 88.4 & 92.9 & 97.3 & 97.1 & 93.8 \\ 
            W$_\textnormal{OCR-read}$ w/ SCOB &202M &\checkmark & 93.2 & 88.8 & 93.7 &97.0  & 96.6& 93.9 \\
            W$_\textnormal{text-read}$ &202M &\checkmark & 93.2 & 88.8 & 92.4 & 97.1 & 96.7 & 93.6 \\ 
            W$_\textnormal{text-read}$ w/ SCOB &202M &\checkmark & 93.5 & 89.1 & 93.0 &96.9  & 96.8& 93.9 \\
            \bottomrule
        \end{tabular}}
    \caption{The public benchmark on PubLayNet~\cite{zhong2019publaynet} validation set (mAP @ IOU [0.50:0.95]) for document layout analysis. \textit{Abbr.}~\textit{G}: generation model.} \vspace{-0.5em}
    \label{suptab:layout_analysis}
     \end{center}
\end{table}
}

Table~\ref{suptab:layout_analysis} shows the performance of PubLayNet~\cite{zhong2019publaynet}. 
Our models demonstrate comparable performance to previous methods. Importantly, our models can handle multiple downstream tasks using a single pipeline, namely the sequence generation framework, which distinguishes them from previous methods that require specific architectural designs for each task.

\subsection{Comparison on OCR} 
Table~\ref{suptab:ocr_details} provides the F-measure scores for the end-to-end OCR benchmark on scene text. Our model achieves state-of-the-art performance in IC13 and shows comparable results in IC15 and Total-Text compared to the alternative state-of-the-art methods.

\subsubsection{W$_{\textnormal{OCR-read}}$ vs. SPTS}
Our W$_{\textnormal{OCR-read}}$ architecture is similar to SPTS~\cite{peng2022spts}, but outperforms it significantly. Notably, W$_{\textnormal{OCR-read}}$ and SPTS differ in dataset, encoder backbone, and resolution. SPTS pre-trains on Curved Synthetic Dataset 150K~\cite{liu2020abcnet}, MLT-2017~\cite{nayef2017icdar2017}, ICDAR2013~\cite{karatzas2013icdar}, ICDAR2015~\cite{karatzas2015icdar}, and TotalText~\cite{ch2017total}, followed by fine-tuning on each target dataset. In contrast, W$_{\textnormal{OCR-read}}$ generalizes not only to OCR but also to diverse downstream tasks, using various datasets of VDU and STU. While SPTS employs ResNet-50 and Transformer 6 layers as its encoder backbone, we use the Swin-transformer. Moreover, SPTS adopts 1600 resolution, while W$_{\textnormal{OCR-read}}$ uses 768 resolution in pre-training and 2560 resolution in fine-tuning. Notably, the resolution has a significant impact on OCR performance. We also accelerate end-to-end recognition convergence using the auxiliary loss~\cite{carion2020end}. Unlike SPTS, we do not use word instance padding, allowing our model to learn many more words with the same decoder sequence length by saving redundant decoding sequences. While word instance padding may help the model converge at the early training stages, it does not significantly improve final performance.

\begin{figure*}[t]
\begin{center}
   \includegraphics[width=0.9\textwidth]{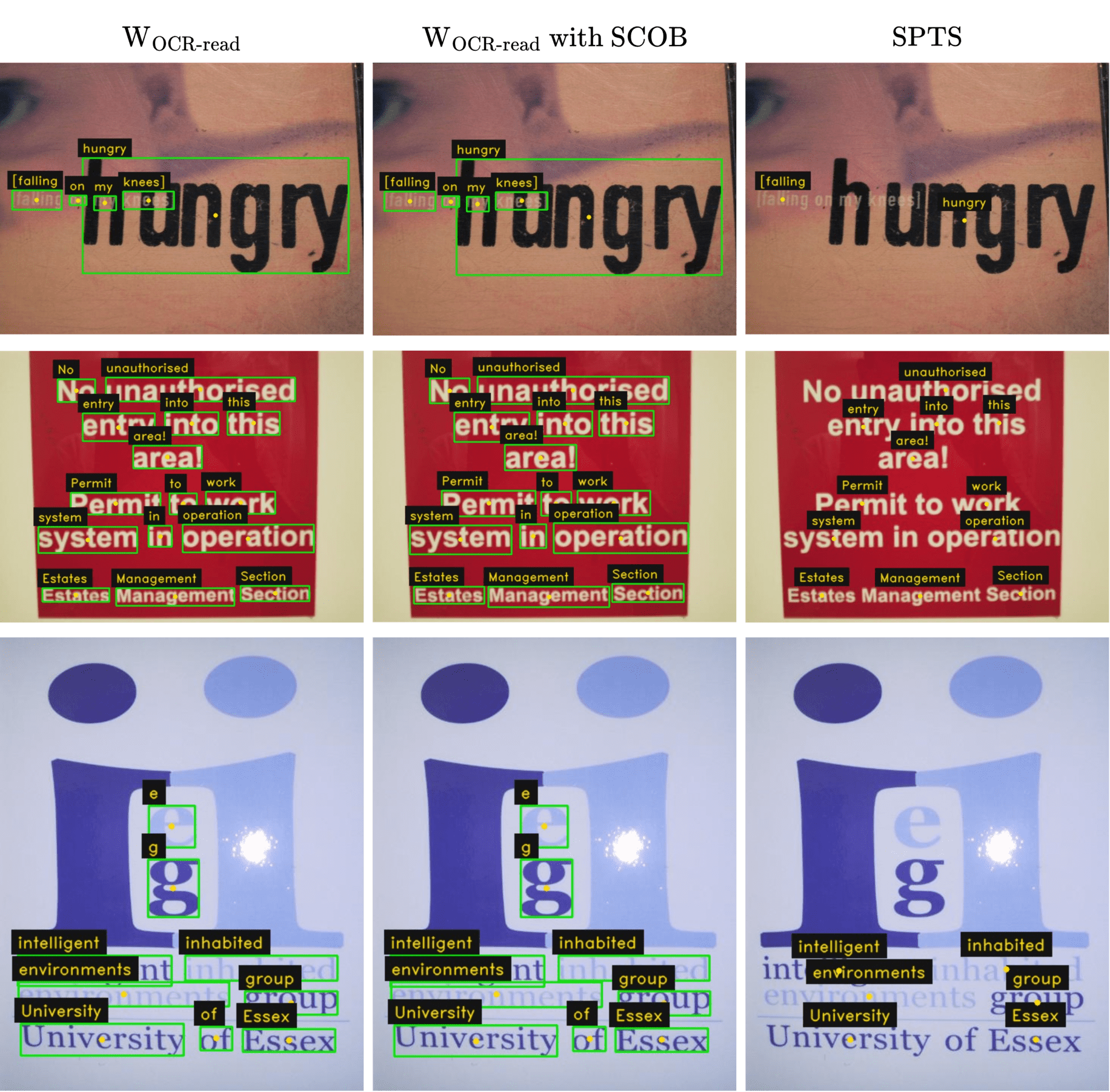}
\end{center}
   \caption{
   The visualization of OCR prediction on ICDAR2013 test set. Our W models predicts in the form of a bounding box and SPTS predicts in the form of single-points. Note that the central points of the bounding boxes predicted by our W models are displayed for comparison.
   } 
   \label{subfig:ic13}
\end{figure*}

\begin{figure*}[t]
\begin{center}
   \includegraphics[width=0.9\textwidth]{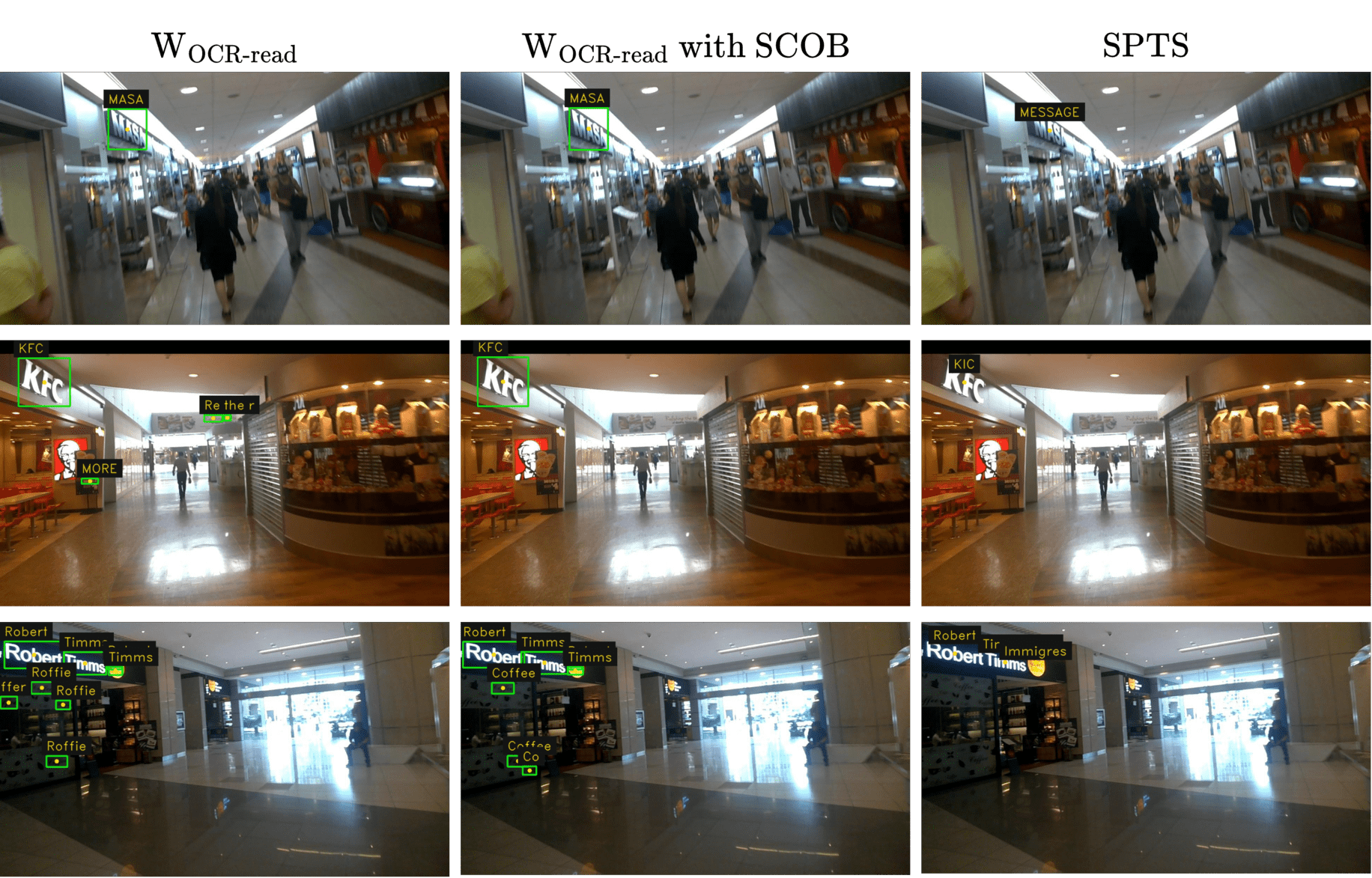}
\end{center}
   \caption{
   The visualization of OCR prediction on ICDAR2015 test set. Our W models predicts in the form of a bounding box and SPTS predicts in the form of single-points. Note that the central points of the bounding boxes predicted by our W models are displayed for comparison.
   } 
   \label{subfig:ic15}
\end{figure*}

\begin{figure*}[t]
\begin{center}
   \includegraphics[width=0.9\textwidth]{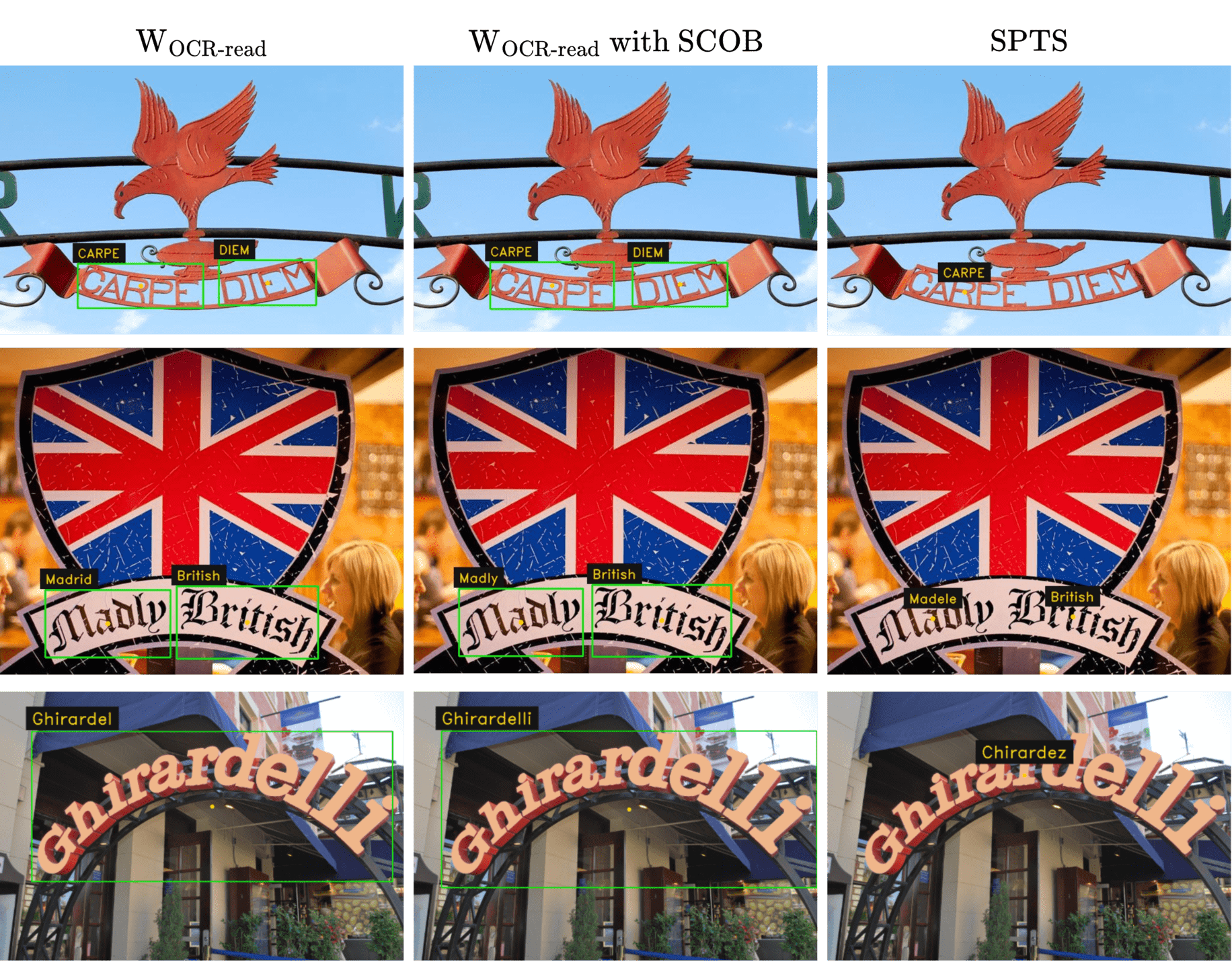}
\end{center}
   \caption{
   The visualization of OCR prediction on TotalText test set. Our W models predicts in the form of a bounding box and SPTS predicts in the form of single-point. Note that the central points of the bounding boxes predicted by our W models are displayed for comparison.
   } 
   \label{subfig:totaltext}
\end{figure*}

To compare two methods that have different outputs for coordinate information, the central point of the box of our model is taken as a single point, and we follow the SPTS evaluation protocol.
Figures~\ref{subfig:ic13}-\ref{subfig:totaltext} also present the qualitative comparisons: W$_\textnormal{OCR-read}$, W$_\textnormal{OCR-read}$ with SCOB, and SPTS. As can be seen, W is more robust to small or dense words. As shown in Figure~\ref{subfig:totaltext}, W also predicts well for curved text.

\subsection{Comparison with LayoutLMv3}
As shown in Table~\ref{suptab:layoutlmv3}, we present the performance of LayoutLMv3 based on different OCR models.
When employing lightweight OCR models like EasyOCR\footnote{\url{https://github.com/JaidedAI/EasyOCR}} and PaddleOCR\footnote{\url{https://github.com/PaddlePaddle/PaddleOCR}}, LayoutLMv3 demonstrates improved speed compared to our model.
However, it is important to note that the scores obtained with these lightweight OCR models are significantly inferior when compared to the utilization of a commercial OCR model like MSAzure\footnote{\url{https://learn.microsoft.com/en-us/azure/cognitive-services/computer-vision/overview-ocr}}.

\setlength{\tabcolsep}{4pt}
\renewcommand{\arraystretch}{1.3} 
\begin{table}
    \begin{center}  
    \centering
    \resizebox{1.0\linewidth}{!}{
    \begin{tabular}{@{}lcccc} \toprule    
    Models & TEDS & Model Params. &OCR Params. & Time (s/img) \\ \midrule    
    LayoutLMv3 + EasyOCR & 56.2 & 133M &25M & 0.7 \\
    LayoutLMv3 + PaddleOCR & 60.5 & 133M &12M & 0.3 \\
    LayoutLMv3 + MSAzure &  84.4 & 133M &n/a  & 1.8  \\ \midrule   
    \textbf{W}$_\textnormal{OCR-read}$ & 88.2 & 202M & None & 1.1 \\ \bottomrule
    \end{tabular}
    }     
    \caption{The performance and inference time on CORD testset. LayoutLMv3 can be combined with various OCR models.}    
    \label{suptab:layoutlmv3}
    \end{center}
\end{table}

\section{Discussion}
We reported a wide range of benchmark results, some of which may not be desirable for validating SCOB.
This is because we hope to contribute to the text understanding field by presenting a transparent investigation rather than hiding adverse findings.
The sequence generation model inherently suffers from the limitation of maximum sequence length~\cite{kim2021donut, wang2022git}.
For future works, it would be important to solve this problem, which may help the model further learn document understanding.

\myparagraph{Character vs. Subword Tokenizer.}
Despite a subword tokenizer's efficiency, we opted for a character tokenizer to improve performance on both VDU and STU tasks. Notably, using a subword tokenizer led to a drop in OCR performance, while other tasks maintained similar performance. This choice is consistent with OCR models like SPTS and UNITS~\cite{Kil_2023_CVPR}. While a decoding length of 512 may appear concerning, it facilitated larger batch sizes with the same GPU memory. For certain tasks, such as table reconstruction, we increased the decoding length during fine-tuning, thereby enhancing our final performance.

\myparagraph{Random Placement of Words in Synthetic Images.}
In our study, we discerned a consistent stability in SCOB training across both the VDU and STU domains. Interestingly, even though the absence of real data coordinates in SCOB impinges upon the OCR's detection performance, it enhances recognition capabilities and promotes stability in training by leveraging more accessible rendering data. While the absence of word coordinates in real data could potentially disrupt the prediction of subsequent words, we posit that the employment of a teacher-forcing scheme, which feeds in ground-truth words during training, effectively mitigates this issue. Additionally, Sinha~\etal~\cite{sinha-etal-2021-masked} found that word co-occurrence statistics are more crucial than word order in MLM pre-training, which may explain why SCOB can pre-train effectively.

\section{Visualization of t-SNE}
\label{sec:visual}
\begin{figure*}[t]

\begin{center}
   \includegraphics[width=0.9\textwidth]{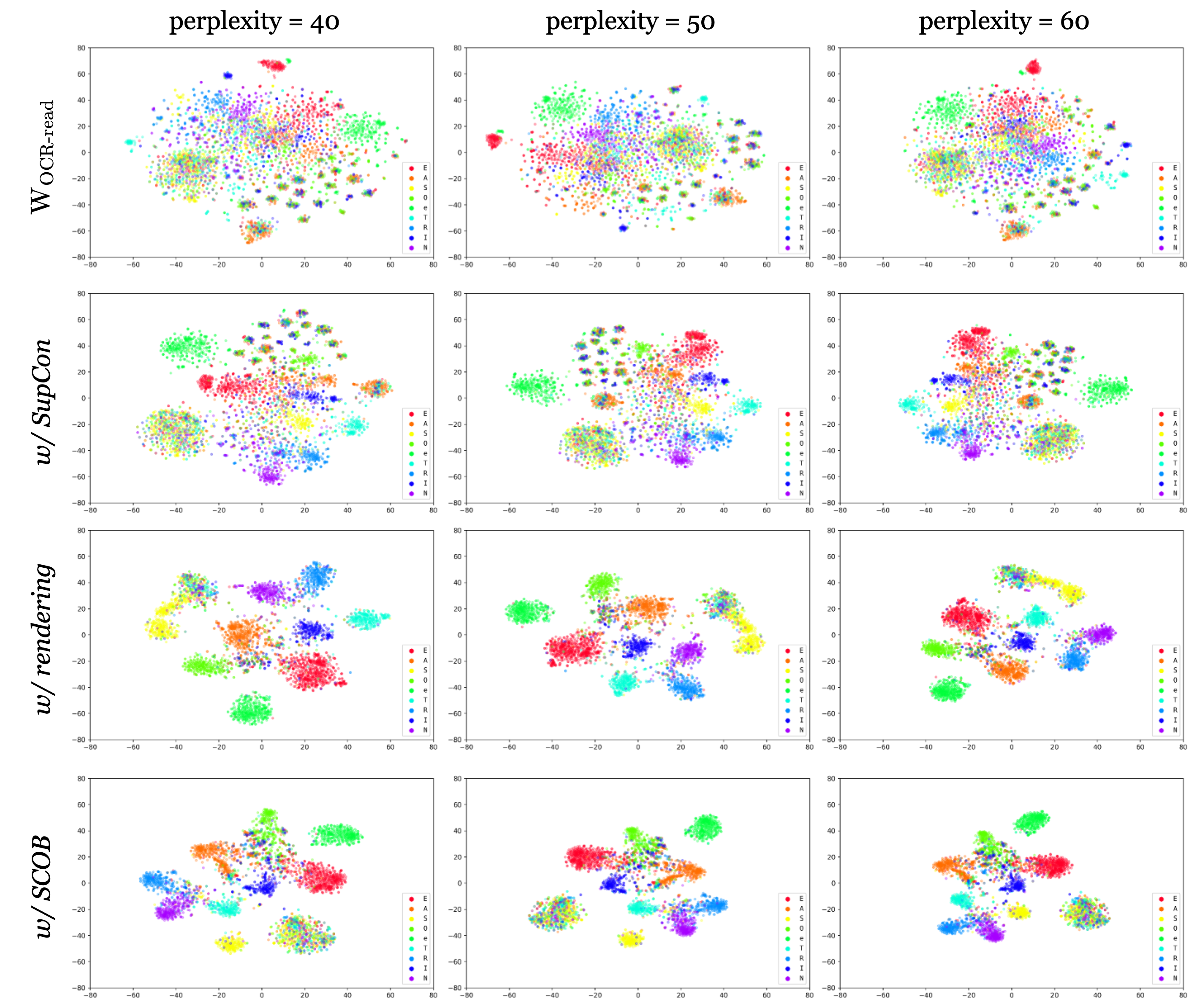}
\end{center}
   \caption{
   Examples of t-SNE visualization. We visualize representations extracted from the final layer of the decoder. Note that the models are pre-trained weights. \textit{Data}: ICDAR2015 test set.
   } 
   \label{subfig:more_tsne_pretrain}
\end{figure*}

\begin{figure*}[t]
\begin{center}
   \includegraphics[width=0.9\textwidth]{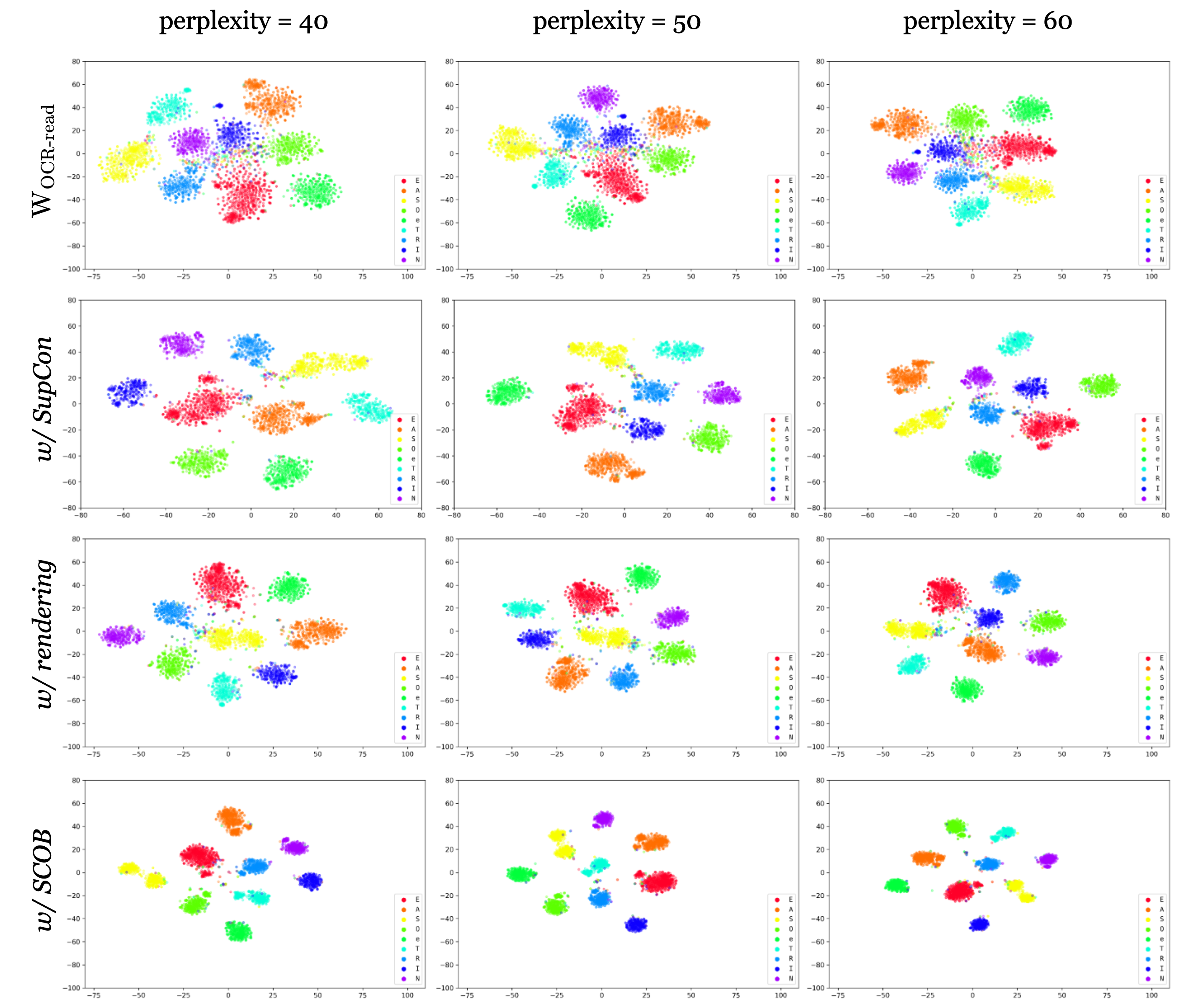}
\end{center}
   \caption{
   Examples of t-SNE visualization. We visualize representations extracted from the final layer of the decoder. The models are fine-tuned for scene text OCR. \textit{Data}: ICDAR2015 test set.
   } 
   \label{subfig:more_tsne}
\end{figure*}

Figures~\ref{subfig:more_tsne_pretrain} and~\ref{subfig:more_tsne} provide more diverse results visualized by t-SNE~\cite{tsne2008} with various perplexities. 
To color-code points according to ground truth classes in sequence generation architecture, we use a teacher-forcing scheme. 
Note that Figures~\ref{subfig:more_tsne_pretrain} and~\ref{subfig:more_tsne} use pre-trained and fine-tuned models, respectively.

As shown in the figures, our visualizations show similar trends regardless of perplexity values, but visualization for pre-trained models have a different tendency with fine-tuned models.
The ICDAR2015 dataset contains small and blurry texts that can be inaccurately identified in low-resolution images. 
Considering models are pre-trained with small resolution ($768\times768$) images, all models are inherently incapable of identifying very small and blurry texts.
These texts might be visualized as multi-colored clusters composed of different classes in Figure~\ref{subfig:more_tsne_pretrain}.
In W$_\textnormal{OCR-read}$ with SCOB and W$_\textnormal{OCR-read}$ with online text rendering, a large multi-colored cluster remains, but scattered small multi-colored clusters disappear. This could be because our online text renderer makes the model robust against small and blurry texts by rendering texts with various image augmentations. As can be seen in Figure~\ref{subfig:more_tsne}, our W$_\textnormal{OCR-read}$ with SCOB extracts the latent representations more discriminatively than other models in the embedding space.

\end{document}